\documentclass[letterpaper, 10 pt, conference]{ieeeconf}  %

\IEEEoverridecommandlockouts{}                            %

\usepackage{amsmath} %
\usepackage{amssymb}  %

\usepackage{graphicx} %
\usepackage{algorithm,algorithmic} %
\usepackage{subfigure} %
\usepackage{color} %
\usepackage{csquotes} %
\usepackage{xcolor} %
\usepackage{siunitx} %
\usepackage{balance}

\graphicspath{{Media/}}

\newtheorem{defn}{Definition}[section]

\usepackage{eso-pic}
\newcommand{\mycopyrighttext}{%
  \footnotesize
  \noindent
  \textcopyright~2026 IEEE. Personal use of this material is permitted.
  Permission from IEEE must be obtained for all other uses, in any current
  or future media, including reprinting/republishing this material for
  advertising or promotional purposes, creating new collective works,
  for resale or redistribution to servers or lists, or reuse of any
  copyrighted component of this work in other works.\\
  IEEE International Conference on Robotics and Automation (ICRA) - June 1-5, 2026.
}

\AtBeginDocument{%
  \AddToShipoutPicture*{%
    \AtPageUpperLeft{%
      \put(55, -40){%
        \parbox{\dimexpr\textwidth-4pt\relax}{\raggedright \mycopyrighttext}
      }%
    }
  }
}

\title{\LARGE \bf
Constraint Manifold Exploration for Efficient Continuous\\
Coverage Estimation
}

\author{Robert Wilbrandt\(^{1}\) and Rüdiger Dillmann\(^{1}\)%
\thanks{\(^{1}\)FZI Research Center for Information Technology, Haid-und-Neu-Straße 10--14,
        76131 Karlsruhe, Germany
        \texttt{\{wilbrandt,dillmann}@fzi.de\}}
}

\begin{document}

\maketitle
\thispagestyle{empty}
\pagestyle{empty}

\begin{abstract}
Many automated manufacturing processes rely on industrial robot arms to move process-specific tools along workpiece surfaces.
In applications like grinding, sanding, spray painting, or inspection, they need to cover a workpiece fully while keeping their tools perpendicular to its surface.
While there are approaches to generate trajectories for these applications, there are no sufficient methods for analyzing the feasibility of full surface coverage.
This work proposes a sampling-based approach for continuous coverage estimation that explores reachable surface regions in the configuration space.
We define an extended ambient configuration space that allows for the representation of tool position and orientation constraints.
A continuation-based approach is used to explore it using two different sampling strategies.
A thorough evaluation across different kinematics and environments analyzes their runtime and efficiency.
This validates our ability to accurately and efficiently calculate surface coverage for complex surfaces in complicated environments.
\end{abstract}

\section{INTRODUCTION}%
\label{sec:introduction}
Industrial robot arms are a vital part of modern automated manufacturing operations.
An extensive range of tasks requires them to move process-specific tools along predefined workpiece surfaces.
The chosen trajectory directly influences the process result: Sanding and polishing rely on uniform tool engagement, spray painting requires a consistent coating, and automated inspection should move continuously to ensure efficient cycle times~\cite{glorieux_coverage_2020}.
All of these tasks must not only cover the entire surface but also keep the tool orthogonal to the surface.

\begin{figure}[t]
    \centering
    \includegraphics[width=.9\columnwidth]{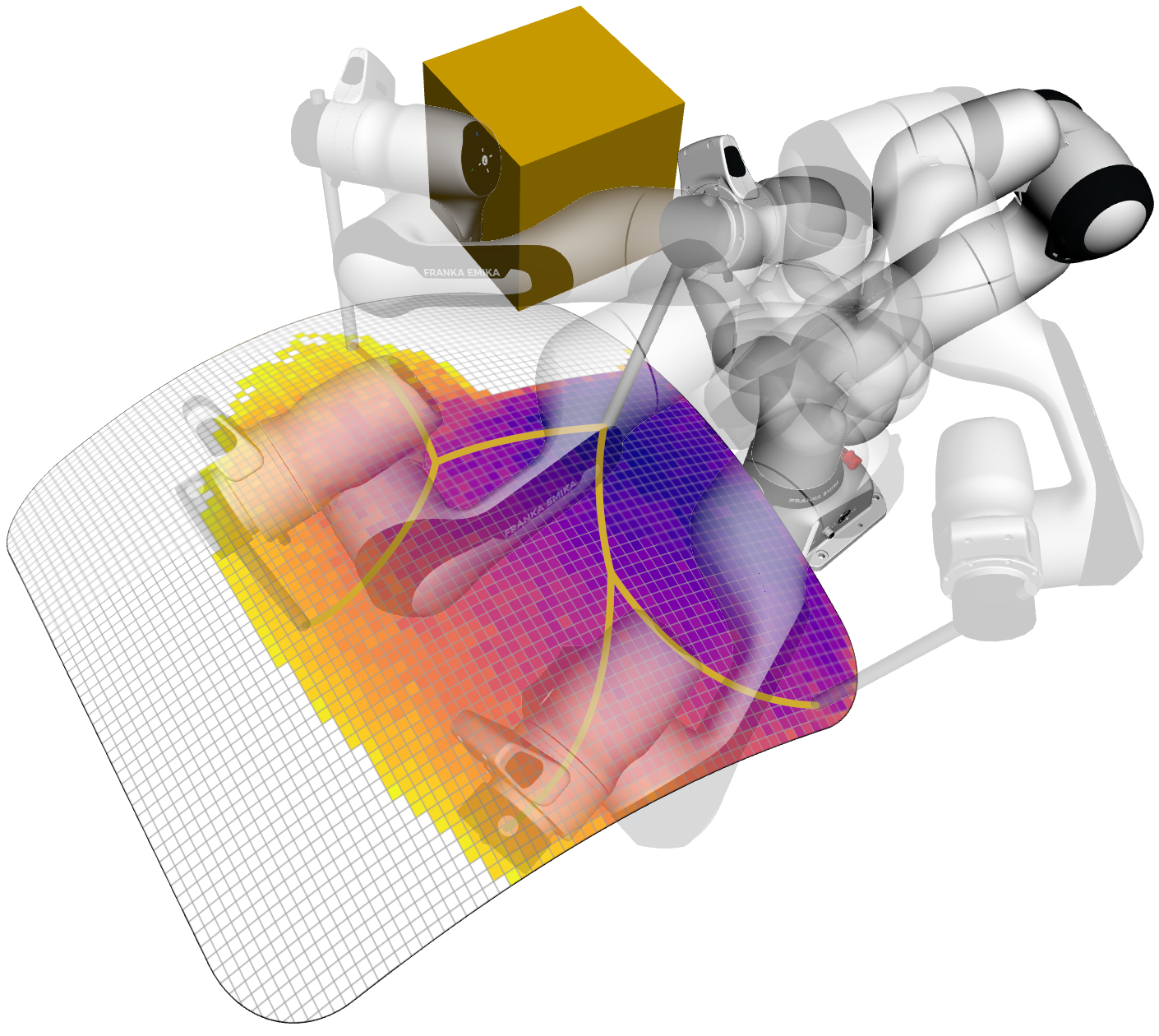}%
    \label{fig:introduction_coverage}
    \caption{Continuous coverage of a surface by a robot using the proposed approach. Starting from an initial configuration, continuous paths across the surface are explored. Reachable surface cells are colored based on their exploration order and some configurations near the border are overlaid.}
\end{figure}

\begin{figure*}
    \centering
        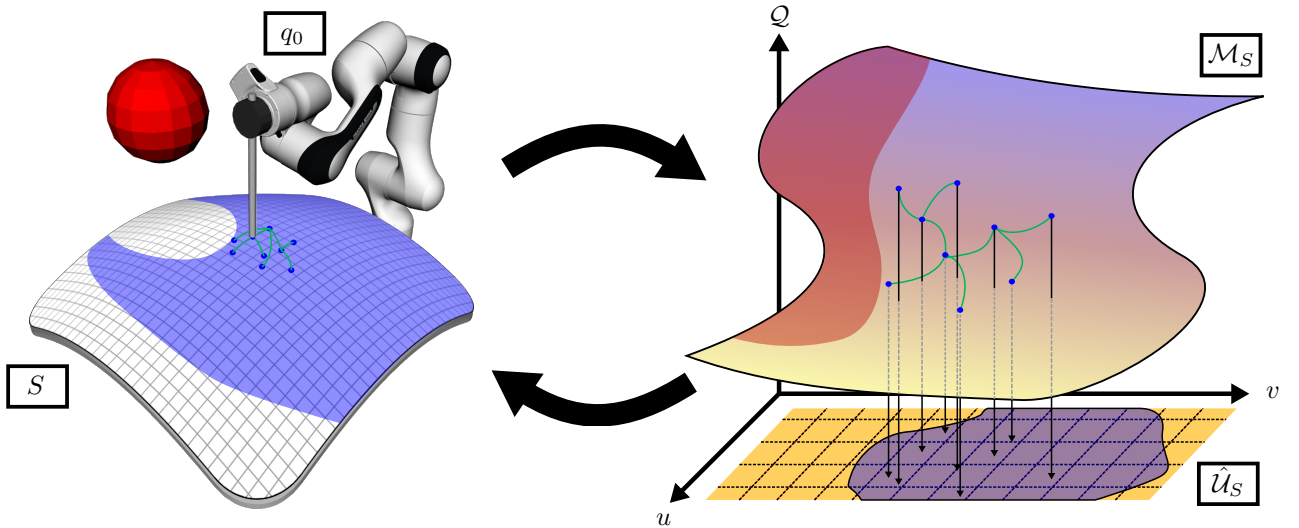 %
    \label{fig:approach_estimation}
    \caption{Continuous coverage estimation by configuration space manifold exploration. Continuous motions of the robot on the left along the surface \(S\) form a smooth manifold \(\mathcal{M}_S\) on the right. Starting from an initial configuration \(q_0\), we explore the manifold using the configuration-space motions shown in green and project them to surface coordinates. Continually doing so increases the estimated coverable region \(\hat{\mathcal{U}}_S\) until the blue region on the left is covered by samples.}
\end{figure*}

There is a significant body of work on planning trajectories for these applications as constrained \emph{Coverage Path Planning} (CPP) problems.
In surface finishing, several approaches plan for contact areas between tool and surface~\cite{wen_uniform_2022} and incorporate segmentation for complex parts~\cite{schneyer_segmentation_2023}.
Automated dimensional inspection requires smooth trajectories to optimize accuracy~\cite{li_five-axis_2023}.
In automated depowdering, covering paths need to address highly complex geometries efficiently~\cite{do_geometry-aware_2023}.
For general coverage planning, local parametrizations of surface geometries were proposed~\cite{lin_robot_2017}, and UV grids for specific types of surfaces can be calculated~\cite{mcgovern_uv_2022}.
Projection methods can apply 2D geometry to complex surfaces~\cite{jafari_surface_2020}, and graph solvers can prevent redundant coverage~\cite{yang_improved_2025}.
While these approaches generally try to cover the entire workpiece geometry, they cannot determine the feasibility of covering a given surface for a specific scenario.

The general problem of trajectory planning with constraints has received significant attention in the literature~\cite{kingston_sampling-based_2018}.
In addition to being able to plan motions along such surfaces, they can incorporate robot dynamics~\cite{bordalba_randomized_2021} or handle changing constraints across the environment~\cite{kingston_informing_2020}.
They are, however, only able to plan single motions and cannot give any indication of coverage feasibility.

Closely related to coverage feasibility, reachability analysis is a well-known robotic problem.
Reachability maps and inverse reachability maps can be used for placement optimization~\cite{makhal_reuleaux_2018}, and advanced maps can analyse nullspace motion in redundant manipulators~\cite{yao_enhanced_2024}.
The generation of reachability maps can be optimized using \(SE(3)\) convolutions~\cite{han_efficient_2021}, and dimensionality reductions can be used to reduce computation and storage requirements for common robot kinematics significantly.
All of these approaches can help determine if and how a pose on a surface is reachable.
They can, however, not decide if this is possible in a continuous motion on the surface without breaking the tool constraints.

There is a prior approach for estimating continuous surface coverage~\cite{mcgovern_efficient_2021} that was subsequently extended to consider the robot jacobian for improved exploration performance~\cite{mcgovern_general_2024}.
While their approach is highly efficient, it relies on discretizations of the surface and the configuration space and thus cannot express constraints exactly.
We also argue in Section~\ref{sec:results} that their approach can underestimate coverage in the presence of obstacles.

This work proposes a novel approach for robot continuous coverage estimation based on randomized sampling of implicit constraint manifolds that can be seen in Figure~\ref{fig:introduction_coverage}.
It can work on complex free-form surfaces and adhere exactly to tool constraints to obtain accurate coverage results.

The remainder of this paper is organized as follows.
Section~\ref{sec:problem_formulation} formally defines the continuous coverage feasibility problem.
In Section~\ref{sec:approach}, we present an approach based on the exploration of an implicit constraint manifold and introduce two exploration strategies.
We compare them across a range of different kinematics and scenarios and analyze their performance in Section~\ref{sec:results}.
Finally, we discuss straightforward extensions to the approach, practical limitations, and avenues for future research in Section~\ref{sec:conclusion}.

\section{PROBLEM FORMULATION}%
\label{sec:problem_formulation}
Consider a robot system with an \(n\)-dimensional configuration space \(\mathcal{Q}\) with \(\mathcal{Q}_{\mathrm{free}} \subseteq \mathcal{Q}\) being the set of non-colliding configurations.
For a given configuration \(q \in \mathcal{Q}\), we refer to \(f_{\mathrm{pos}}(q) \in \mathbb{R}^3\) as the tool position and \(f_{\mathrm{rot}}(q) \in \mathbb{R}^{3\times3}\) as the rotation matrix representation of the robot tool orientation.
By convention and without loss of generality, the orientation of the tool axis is assumed to be given by its z-axis \(f_{\mathrm{rot}}(q) \cdot \begin{pmatrix}
0 & 0 & 1
\end{pmatrix}^T\).

In addition to the robot, we need to consider the surface \(S\) to be evaluated.
Without constraints on a specific representation, we assume the surface to be a \(C^2\)-smooth function \(S: \mathbb{R}^2 \to \mathbb{R}^3\) and only consider a subset of the surface domain \(\mathcal{U}_S \subseteq [u_{\mathrm{min}}, u_{\mathrm{max}}] \times [v_{\mathrm{min}}, v_{\mathrm{max}}]\). %
We write \(n_S(u, v) = \frac{\delta S}{\delta u} \times \frac{\delta S}{\delta v}\) to refer to the surface normal.

When estimating the coverage of the surface, only configurations with tool positions on \(S\) and tool axes orthogonal to \(S\) should be considered, leading to the definition

\begin{defn}[Surface-Constrained Configuration]
    We consider a configuration \(q \in \mathcal{Q}\) \emph{surface-constrained} with regards to a surface \(S\) if there is a surface coordinate \((u, v) \in \mathcal{U}_S\) with \(f_{\mathrm{pos}}(q) = S(u, v)\) and \((f_{\mathrm{rot}}(q) {\begin{pmatrix}0 & 0 & 1\end{pmatrix}}^T) \cdot \hat{n}_S(u, v) = 1\).
\end{defn}

We write \(\mathcal{Q}_S \subseteq \mathcal{Q}\) to refer to the set of all surface-constrained configurations and \(\mathcal{Q}_S^\star = \mathcal{Q}_S \cap \mathcal{Q}_{\mathrm{free}}\) for the subset that is not in collision.

For robot trajectories \(\sigma: [0, 1] \to \mathcal{Q}\) where each \(\sigma(t)\) is surface-constrained with regards to \(S\), there is a surface path \(\sigma_S: [0, 1] \to \mathcal{U}\) such that \(\sigma(t)\) satisfies all motion constraints with regards to \(\sigma_S(t)\) for all \(t \in [0, 1]\).
We can use this to define the problem

\begin{defn}[Continuous Coverage Estimation]
    Given an initial configuration \(q_{\mathrm{start}} \in \mathcal{Q}_S^\star{}\), find the set \(\hat{\mathcal{U}}_S \subseteq \mathcal{U}_S\) of surface coordinates \((u_{\mathrm{cc}}, v_{\mathrm{cc}})\) for which a trajectory \(\sigma^\star: [0, 1] \to \mathcal{Q}_S^{\star}\) exists with  \(\sigma_S^\star(1) = (u_{\mathrm{cc}}, v_{\mathrm{cc}})\).
\end{defn}

\section{APPROACH}%
\label{sec:approach}
We propose a sampling-based approach for estimating continuous coverage, as shown in Figure~\ref{fig:introduction_coverage}.
For this, we define an extended configuration space \(\mathcal{Q}^{+}\) that can directly express the constraints on tool position and orientation.
The resulting implicit manifold directly maps to \(\mathcal{Q}_S\), enabling the exploration of continuously reachable surface coordinates through continuous configuration-space motions.
We present two methods for its exploration.
Both of them perform probabilistic sampling and run continuously, but at any point in time, we can stop and evaluate the explored samples and estimate the continuously covered subset of \(\hat{\mathcal{U}}_S\).

\subsection{Constrained Configuration Space}%
\label{subsec:approach_space}

In order to construct the space of all surface-constrained configurations, we first create an ambient space consisting of robot configurations and surface coordinates.
We can formalize this as \begin{equation}
    \mathcal{Q}^+ = \{{\,}(q, u, v) \mid q \in \mathcal{Q}, (u, v) \in \mathcal{U}{\,}\}
\end{equation}

We want to constrain this space to only configurations \((q, u, v)\) where the robot configuration \(q\) adheres to tool position- and orientation constraints at surface position \((u, v)\).
For both constraints, we define constraint functions that are zero if their constraint is fulfilled.
For the position constraint, we achieve this using \begin{equation}
    C_{\mathrm{pos}} = f_{\mathrm{pos}}(q) - S(u, v)
\end{equation}

For the tool orientation constraint, we consider the surface normal in the coordinate frame of the robot tool.
For constraint-satisfying configurations, this should be colinear with the z-axis.
We can thus define \begin{equation}
    C_{\mathrm{rot}} = \begin{pmatrix}1 & 0 & 0\\ 0 & 1 & 0\end{pmatrix}f_{\mathrm{rot}}^{-1}(q) \cdot \mathbf{n}_S(u, v)
\end{equation}

While this also allows configurations where the robot tool opposes the surface normal, this is not a problem in this application, as a transition to this \emph{flipped} state is impossible.

The combined constraint function \(C = \begin{bmatrix}C_{\mathrm{pos}} & C_{\mathrm{rot}}\end{bmatrix}^T\) implicitly defines the constrained configuration space \begin{equation}
    \mathcal{M} = \{ q^{+} \in \mathcal{Q}^{+} \mid C(q) = \mathbf{0} \}
\end{equation}

By construction, this space is a smooth manifold in \(\mathcal{Q}^+\) and directly maps to \(\mathcal{Q}_S\).
While we generally cannot parametrize it globally, we can explore \(C\)'s nullspace using its jacobian to explore it gradually.
With \(\mathcal{Q}^+\) being of dimension \((n+2)\), \(\mathcal{M}\) should be of dimension \((n-3)\).

\begin{algorithm}[t]
    \caption{RRT Exploration of Constraint Manifold}%
    \label{alg:exploration_rrt}
    \begin{algorithmic}[1]
        \renewcommand{\algorithmicrequire}{\textbf{Input:}}
        \renewcommand{\algorithmicensure}{\textbf{Output:}}
        \REQUIRE{} Initial Robot Configuration \(q_0 \in \mathcal{Q}_{\mathrm{free}}\)
        \STATE{} Find closest point \(S(u_0, v_0)\) to \(f_{\mathrm{pos}}(q_0)\)
        \STATE{} Project \((q_0, u_0, v_0)\) to obtain \(q_S \in \mathcal{M}\)
        \STATE{} Initialize tree \(T\) with root \(q_S\)
        \LOOP{}
            \STATE{} \(q \gets \textsc{SampleUniform}(\mathcal{M})\)
            \STATE{} \(q_N \gets \textsc{NearestNeighbor}(q, T)\)
            \IF{\(\mathrm{dist}(q, q_N) > d_{\mathrm{max}}\)}  %
                \STATE{} \(q \gets \textsc{GeodesicInterpolate}(q_N, q, \frac{d_{\mathrm{max}}}{\mathrm{dist}(q, q_N)})\)  %
            \ENDIF{}
            \IF{\textsc{checkTransition}\((q_N, q, \delta q_{\mathrm{check}})\)}
                \STATE{} \(T \rightarrow \textsc{AddChild}(q, q_N)\)
            \ENDIF{}
        \ENDLOOP{}
    \end{algorithmic}
\end{algorithm}

\subsection{Constraint Manifold Exploration}%
\label{subsec:approach_exploration}

For any given configuration in \(M\), the kernel of the jacobian of \(C\) can be used to construct a local tangent space that acts as a linear approximation in a small neighborhood.
New valid configurations can be found by sampling in this hyperplane and projecting back to \(M\).
We utilize the IMACS~\cite{kingston_exploring_2019} framework for this, which interleaves the sampling of new configurations with the creation of new hyperplanes.
In this context, these tangent spaces are referred to as \emph{charts} and the set of all charts as \emph{atlas}.
IMACS allows us to sample new configurations from a given atlas, project configurations to \(\mathcal{M}\), and interpolate between states along geodesics.
We refer to Kingston et al.~\cite{kingston_exploring_2019} for a detailed description.

Building on IMACS, we propose two algorithms for manifold exploration.
First, we create an RRT that utilizes uniform sampling over all previously covered charts.
This acts as a solid baseline and can provide asymptotic coverage guarantees, but can be inefficient without any sample biasing.
To address this, we introduce a second algorithm based on the KPIECE~\cite{sucan_sampling-based_2012} planner that guides exploration towards unknown regions of the \(S\).

\subsubsection{RRT-Based Exploration}%
\label{subsubsec:approach_exploration_rrt}

Our first method utilizes uniform sampling over the covered atlas to gradually grow the covered region of \(M\) and is shown in Algorithm~\ref{alg:exploration_rrt}.
For its initialization, it first finds the closest surface coordinate \((u_0, v_0)\) to \(f_{\mathrm{pos}}(q_0)\) and then projects \((q_0, u_0, v_0)\) to \(\mathcal{M}\).
The resulting configuration \(q_S \in \mathcal{Q}^+\) is used as the root of the constructed tree \(T\).

The iterative exploration follows the standard RRT procedure.
Using IMACS, we first sample a random configuration \(q\) using the already covered charts, rejecting states outside the joint limits or in collision.
We then find its closest know neighbor \(q_N\) in \(T\) and use the geodesic between \(q_N\) and \(q\) to expand \(T\) in the direction of \(q\).
To ensure continuous motions, we check the transition between them at a regular step of \(\delta q_{\mathrm{check}}\) and make sure that interpolation is possible and does not lead to collisions.

Following the argument by Kingston et al.~\cite{kingston_exploring_2019}, the proof by Jaillet and Porta~\cite{jaillet_path_2013} for the eventual coverage of \(\mathcal{M}\) by the atlas created by IMACS holds.
We therefore argue that for any subset \(\mathcal{U}^{\prime} \subseteq \mathcal{U}\) of non-zero measure that is continuously reachable from \(q_0\), we will eventually sample a configuration \((q, u, v)\) with \((u, v) \in \mathcal{U}^{\prime}\).
Therefore, we are guaranteed to eventually estimate the continuously coverable region exactly using this approach.

\subsubsection{Biased Sampling}%
\label{subsubsec:approach_exploration_biased}

\begin{algorithm}[t]
    \caption{Biased Manifold Exploration}%
    \label{alg:exploration_biased}
    \begin{algorithmic}[1]
        \renewcommand{\algorithmicrequire}{\textbf{Input:}}
        \renewcommand{\algorithmicensure}{\textbf{Output:}}
        \REQUIRE{} Initial Robot Configuration \(q_0 \in \mathcal{Q}_{\mathrm{free}}\)
        \STATE{} Find closest point \(S(u_0, v_0)\) to \(f_{\mathrm{pos}}(q_0)\)
        \STATE{} Project \((q_0, u_0, v_0)\) to obtain \(q_S \in \mathcal{M}\)
        \STATE{} Initialize empty grid \(G\)
        \STATE{} \(G \rightarrow \textsc{Add}(q_S, (u_0, v_0))\)
        \LOOP{}
            \STATE{} \(q \gets \textsc{Select}(G)\)~\label{line:approach_exploration_biased_sample}
            \STATE{} \(q^{\prime} \gets \textsc{SampleGaussian}(\mu=q, \sigma=\sigma_{\mathrm{sample}})\)
            \IF{\(\textsc{CheckTransition}(q, q^{\prime}, \delta q_{\mathrm{check}})\)}
                \STATE{} \(G \rightarrow \textsc{Add}(q^{\prime}, {\left[q\right]}_{uv})\)
            \ENDIF{}
        \ENDLOOP{}
    \end{algorithmic}
\end{algorithm}

While uniform RRT-based exploration has nice theoretical properties and will eventually result in an accurate estimate, it can perform poorly in practice.
This is especially true in cluttered environments with narrow passages, where sampling-based planners traditionally also struggle~\cite{orthey_sampling-based_2024}.

\begin{figure}[t]
    \subfigure{\label{fig:results_scenarios_ur_simple}\includegraphics[width=.32\columnwidth,height=4cm,keepaspectratio]{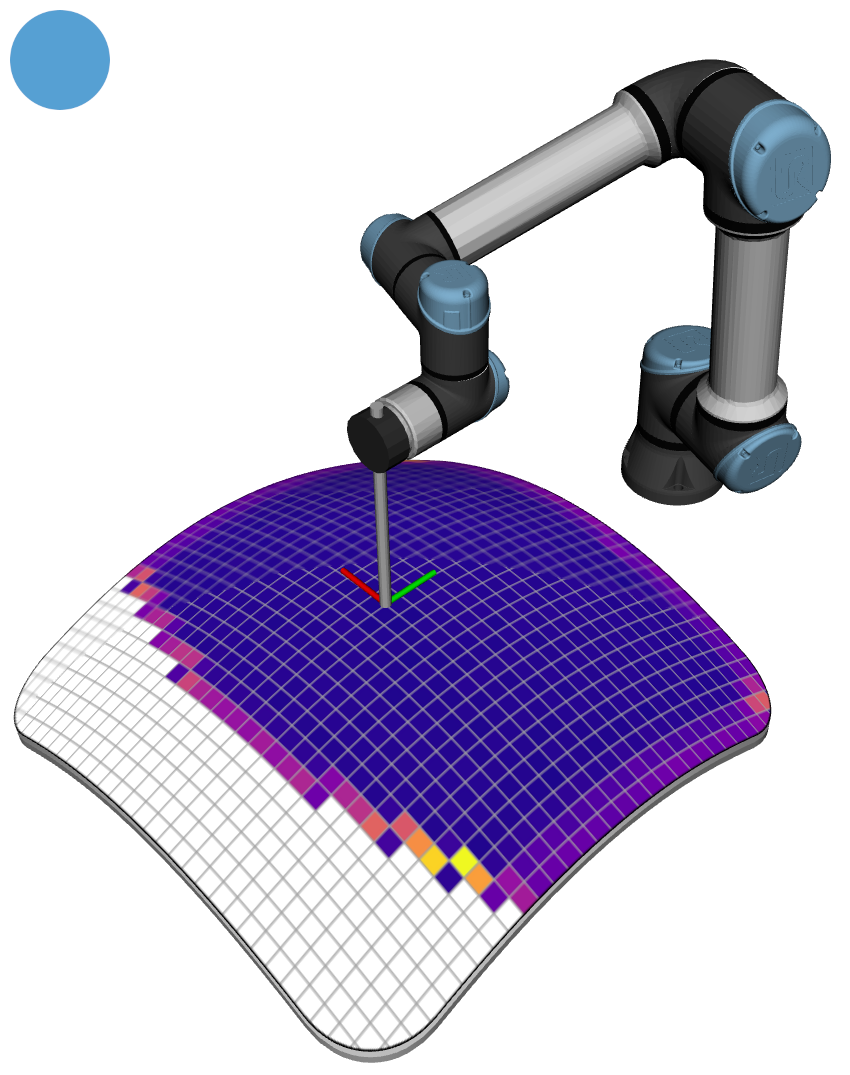}}%
    \hfill
    \subfigure{\label{fig:results_scenarios_kuka_free}\includegraphics[width=.32\columnwidth,height=4cm,keepaspectratio]{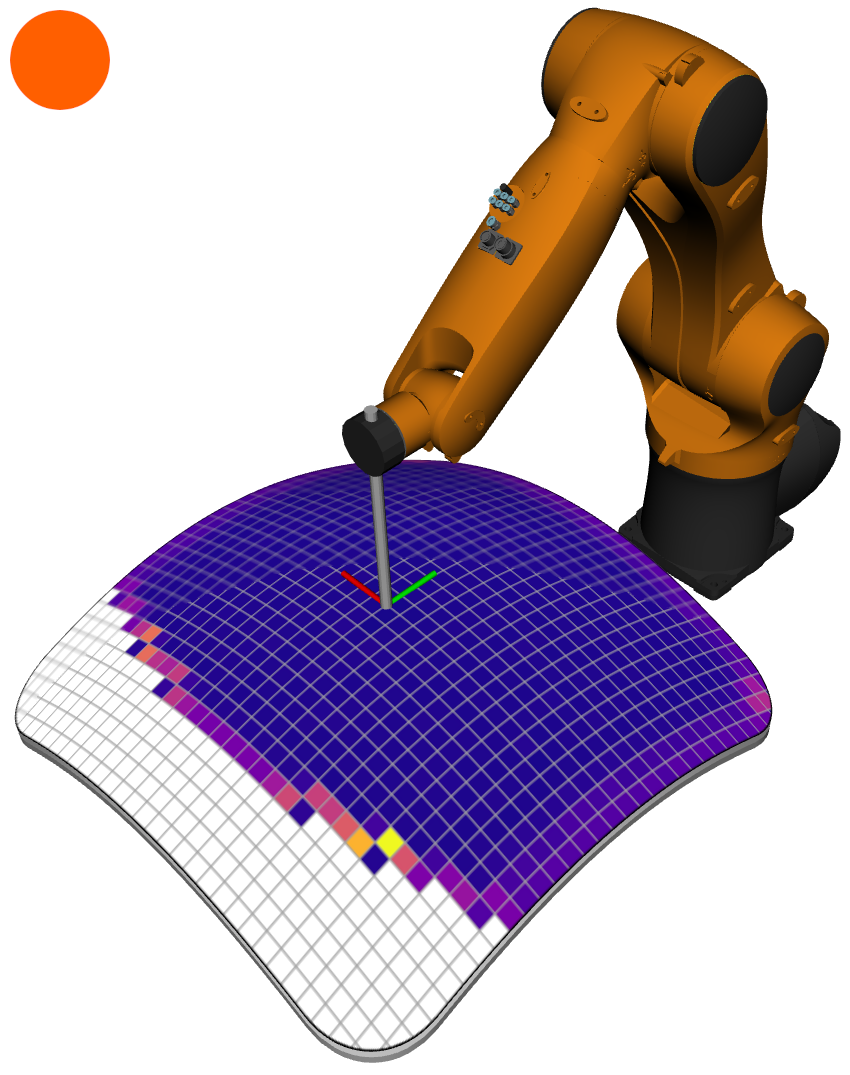}}%
    \hfill
    \subfigure{\label{fig:results_scenarios_franka_free}\includegraphics[width=.32\columnwidth,height=4cm,keepaspectratio]{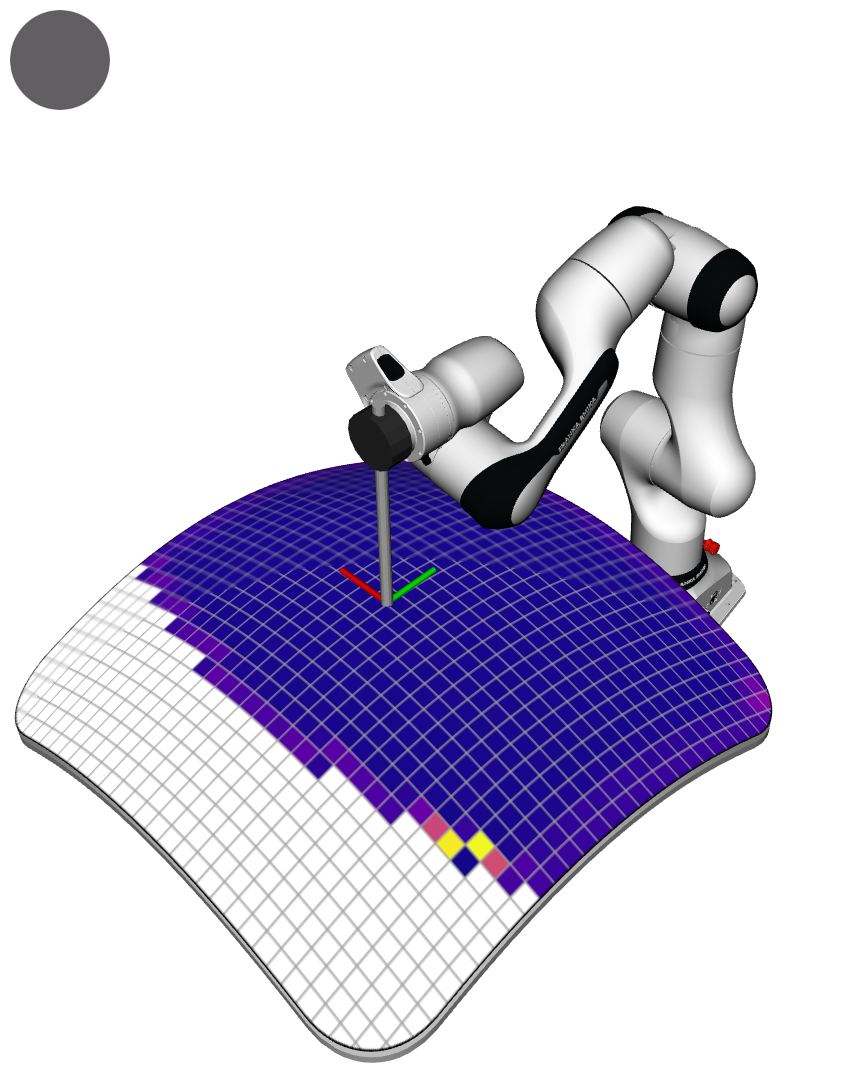}}\
    \subfigure{\label{fig:results_scenarios_ur_cluttered}\includegraphics[width=.32\columnwidth,height=4cm,keepaspectratio]{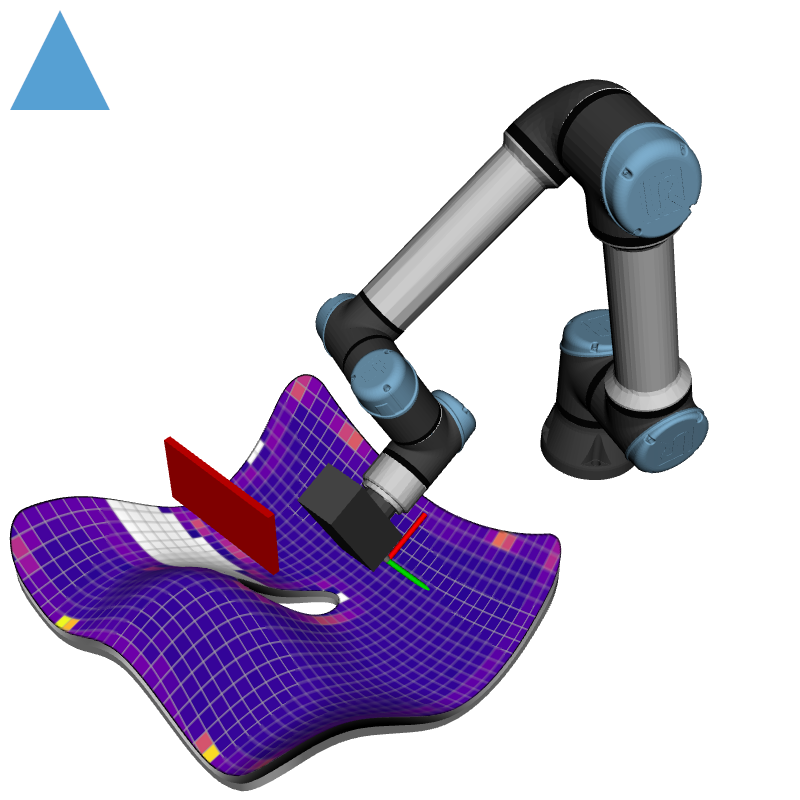}}%
    \hfill
    \subfigure{\label{fig:results_scenarios_kuka_cluttered}\includegraphics[width=.32\columnwidth,height=4cm,keepaspectratio]{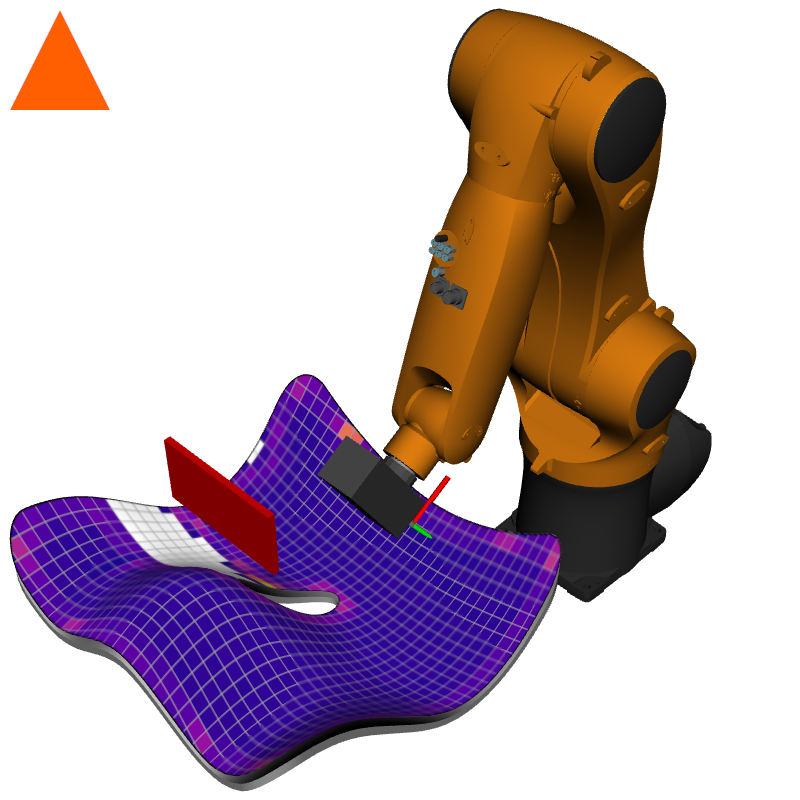}}%
    \hfill
    \subfigure{\label{fig:results_scenarios_franka_cluttered}\includegraphics[width=.32\columnwidth,height=4cm,keepaspectratio]{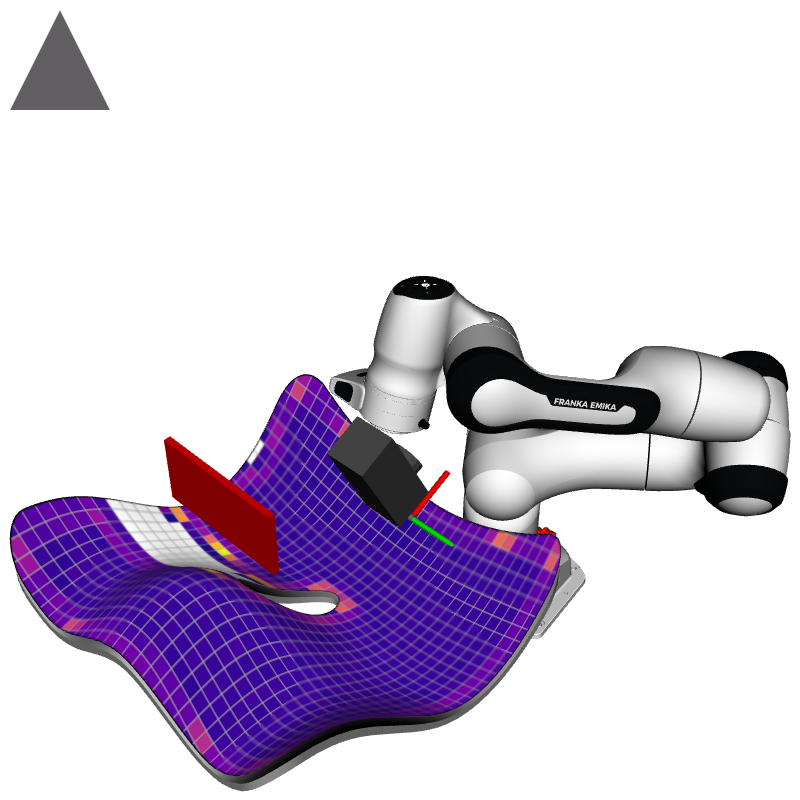}}\
    \subfigure{\label{fig:results_scenarios_ur_maze}\includegraphics[width=.32\columnwidth,height=4cm,keepaspectratio]{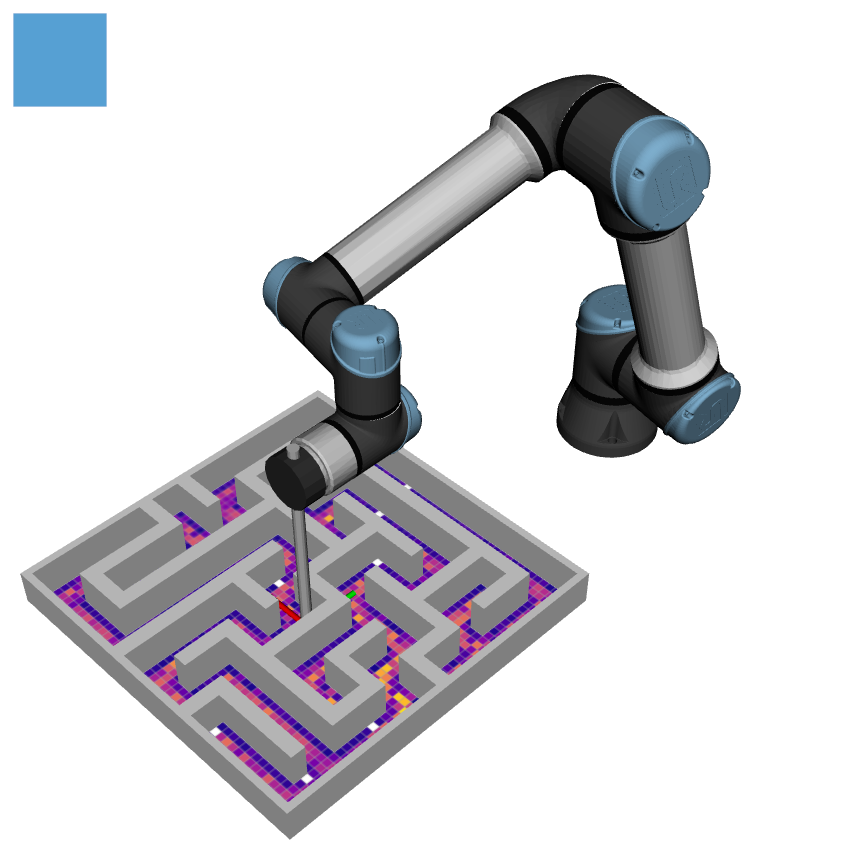}}%
    \hfill
    \subfigure{\label{fig:results_scenarios_kuka_maze}\includegraphics[width=.32\columnwidth,height=4cm,keepaspectratio]{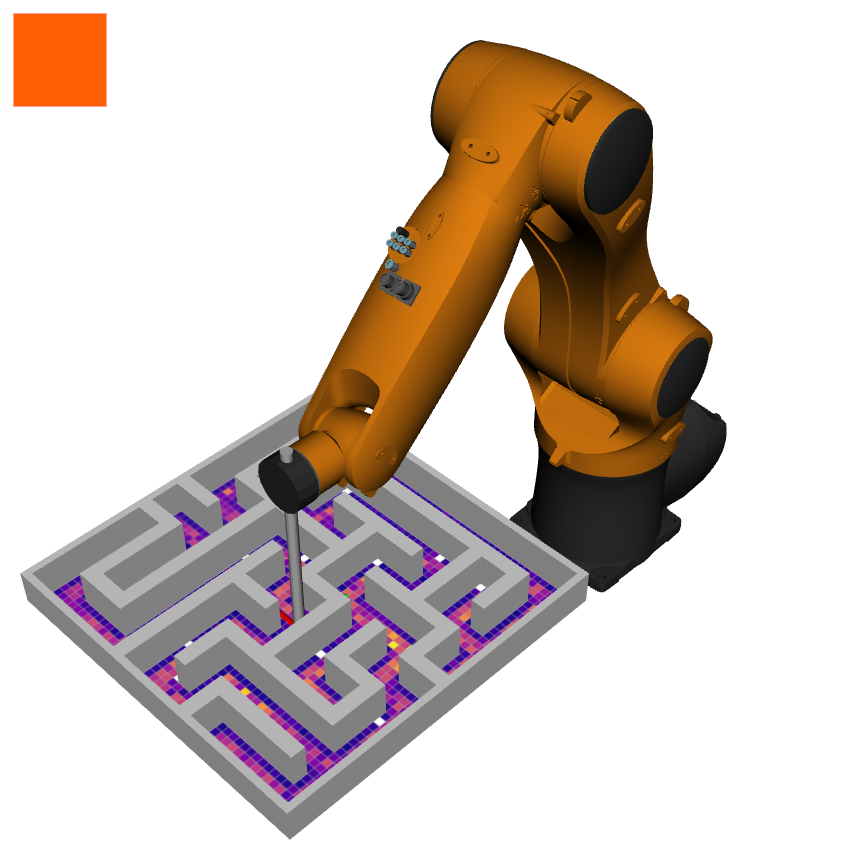}}%
    \hfill
    \subfigure{\label{fig:results_scenarios_franka_maze}\includegraphics[width=.32\columnwidth,height=4cm,keepaspectratio]{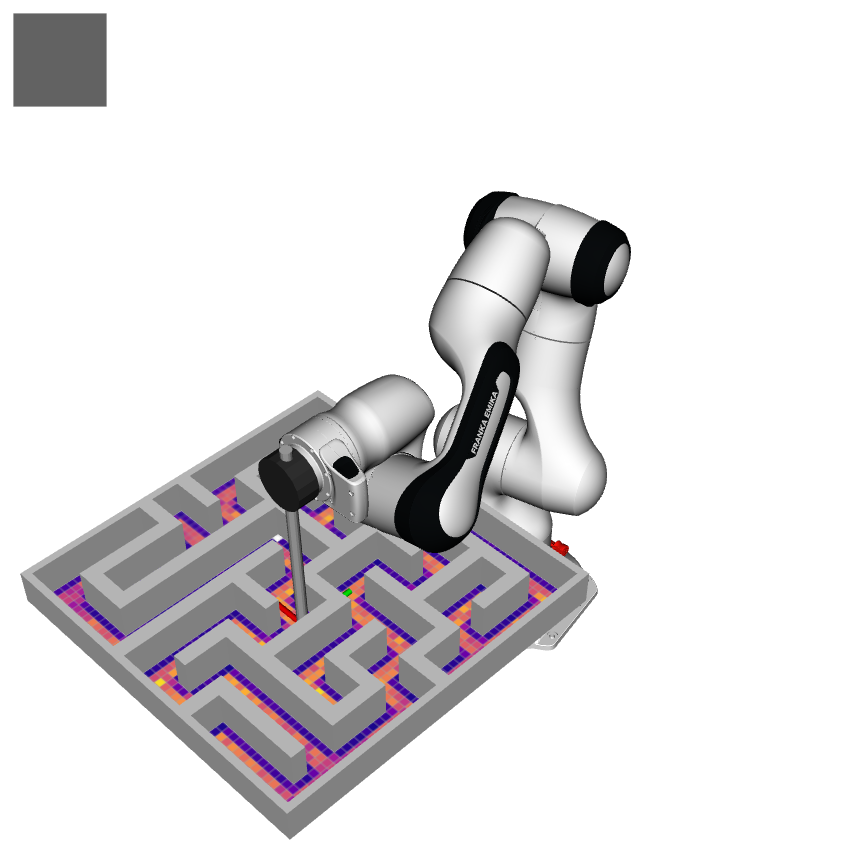}}\
    \caption{The matrix of robots and environments used throughout the evaluation. We use two six-axis and one seven-axis robots to cover different relevant kinematics, and three surfaces with different curvatures and obstacles to achieve representative results. The icons next to each robot are used throughout the evaluation to reference specific setups.}%
    \label{fig:results_scenarios}
\end{figure}

To combat these limitations, we propose a biased sampling exploration scheme based on a 2D projection and the KPIECE~\cite{sucan_sampling-based_2012} sample biasing scheme that can be seen in Algorithm~\ref{alg:exploration_biased}.
It maintains a 2D grid to bias sampling in regions where we expect to get better coverage.
We index the grid based on each configuration \(q \in \mathcal{M}\)'s surface coordinates \({\left[q\right]}_{uv}\) and store all newly sampled configurations in their respective cell.
The grid keeps track of \emph{interior} and \emph{exterior} cells depending on samples in neighboring cells, selecting an \emph{exterior} cell for expansion with a strong bias (\(75\% \) in this work).
Both types of cells are ranked according to the \emph{Importance} of a cell \(z\) as defined by

\begin{equation}\label{eqn:approach_ranking}
    \frac{\log(\mathcal{I}) \cdot score}{\mathcal{S} \cdot (1 + \left|Neighbors(z)\right|) \cdot Coverage(z)}
\end{equation}

with \(\mathcal{I}\) being the iteration at which \(z\) was first visited, \(score\) tracking the progress previously achieved when expanding from it and \(\mathcal{S}\) being the number of times it was previously expanded.
We refer to Sucan et al.~\cite{sucan_open_2012} for a detailed description of the cell selection process.

Just as the RRT-based exploration approach, the initial configuration \(q_0\) is first projected to obtain \(q_S \in \mathcal{M}\).
The grid is then initialized and \(q_S\) is stored in \(G\) at \({\left[q_S\right]}_{uv}\).
Starting in line~\ref{line:approach_exploration_biased_sample}, sampling is now performed by selecting a state \(q\) in \(G\) and sampling around it using a Gaussian distribution to obtain \(q^{\prime}\).
The geodesic between \(q\) and \(q^{\prime}\) is then checked at regular intervals \(\delta q_{\mathrm{check}}\) for joint limit validity and collisions, and added to \(G\) if the transition turned out to be valid.

\subsection{Coverage Estimation}%
\label{subsec:approach_estimation}

After any number of samples, we can analyze all configurations sampled so far to obtain an approximation for the continuously coverable region of \(S\).
For evaluation, we discretize \(\mathcal{U}\) into an equally spaced \(n_{\mathrm{grid}} \times n_{\mathrm{grid}}\) grid and record the number \(v_{i, j}\) of configurations that visited each cell.
We report the set of all cells with \(v_{i, j} \ge 0\) as continuously reachable.
For the biased exploration approach described in Section~\ref{subsubsec:approach_exploration_biased}, we use the same grid size for cell selection as for coverage estimation.

\section{RESULTS AND EXPERIMENTS}%
\label{sec:results}
\begin{figure*}[t]
    \centering
    \includegraphics[width=\textwidth]{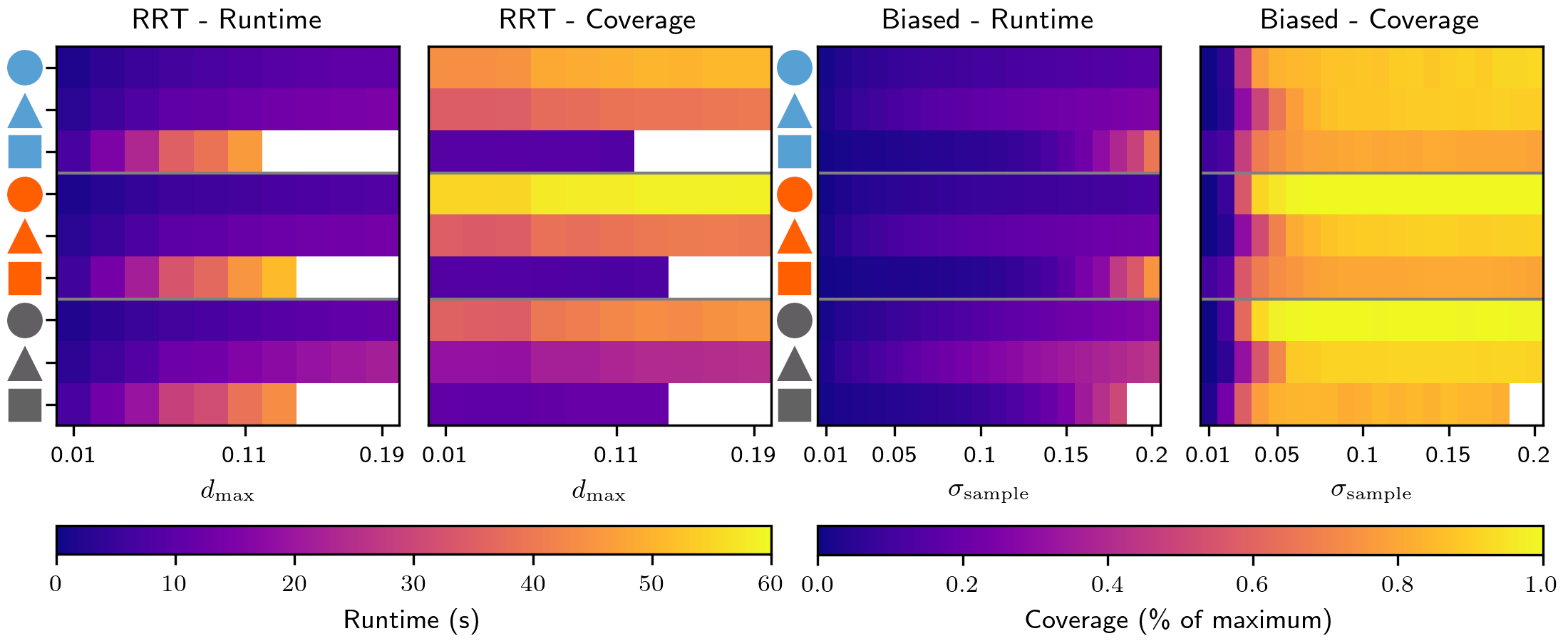}%
    \caption{The results of the parameter evaluation. For both exploration approaches, the mean runtime and the mean coverage in percent of reachable cells across 25 experiments of \num{25000} samples are shown. The rows are organized in groups by scenario.}%
    \label{fig:results_parameters}
\end{figure*}

\begin{figure*}[t]
    \subfigure[RRT]{\label{fig:results_comparison_density_curved_rrt}\includegraphics[width=.12\textwidth]{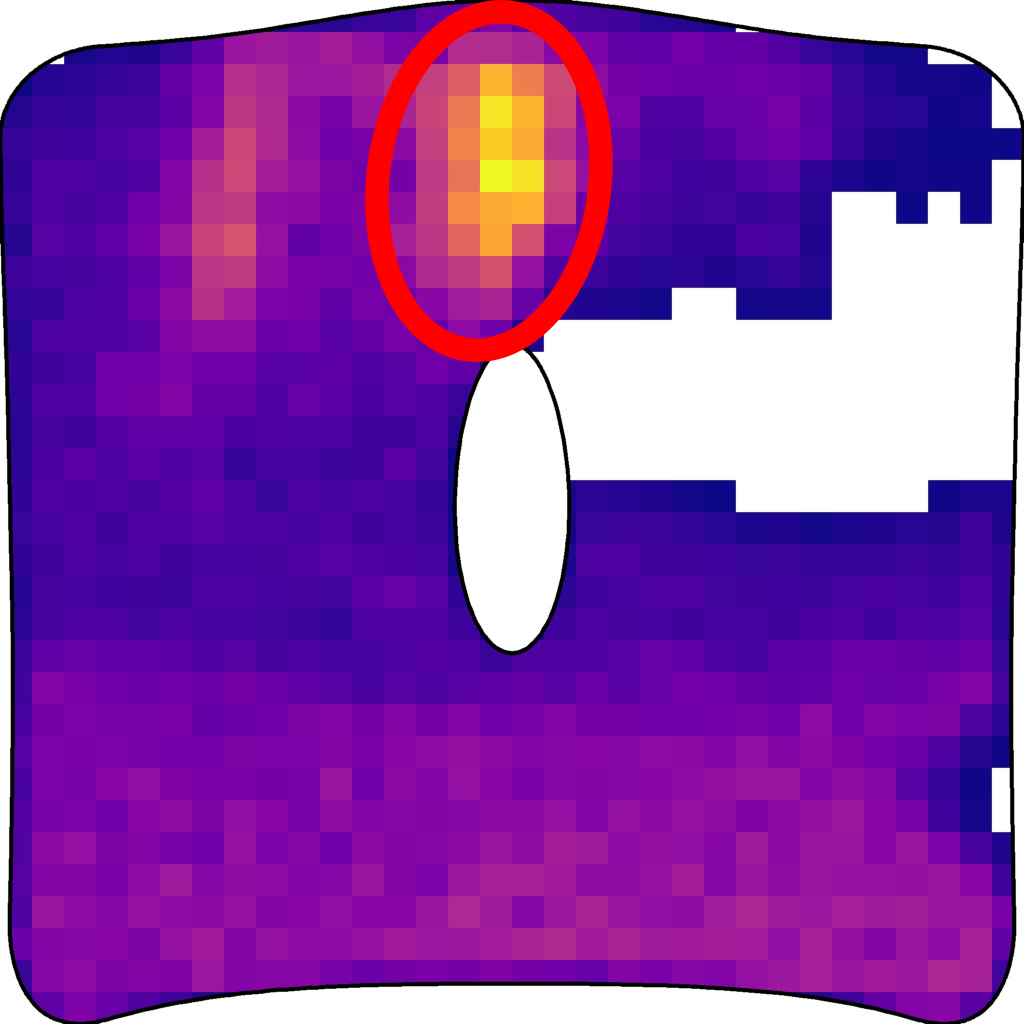}}%
    \hfill
    \subfigure[Biased]{\label{fig:results_comparison_density_curved_biased}\includegraphics[width=.12\textwidth]{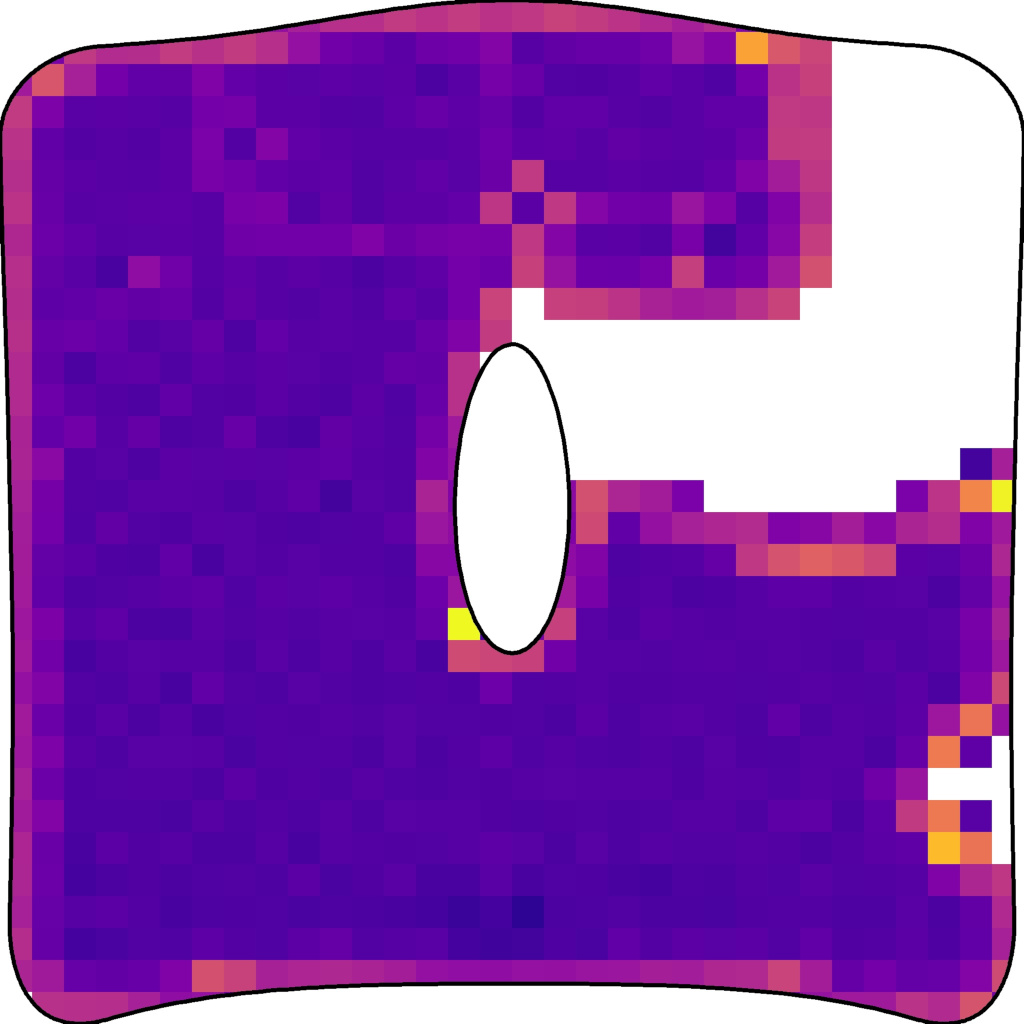}}%
    \hfill
    \subfigure[RRT]{\label{fig:results_comparison_density_maze_rrt}\includegraphics[width=.12\textwidth]{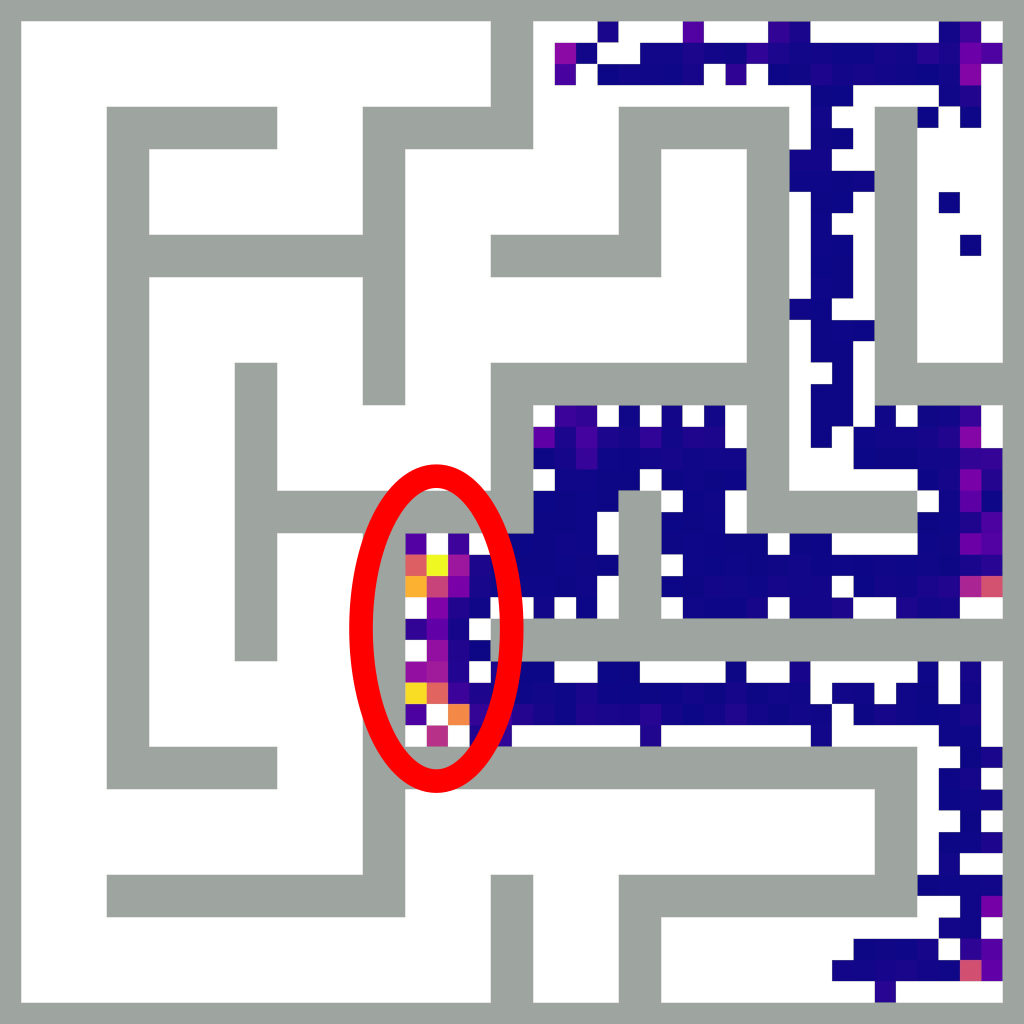}}%
    \hfill
    \subfigure[Biased]{\label{fig:results_comparison_density_maze_biased}\includegraphics[width=.12\textwidth]{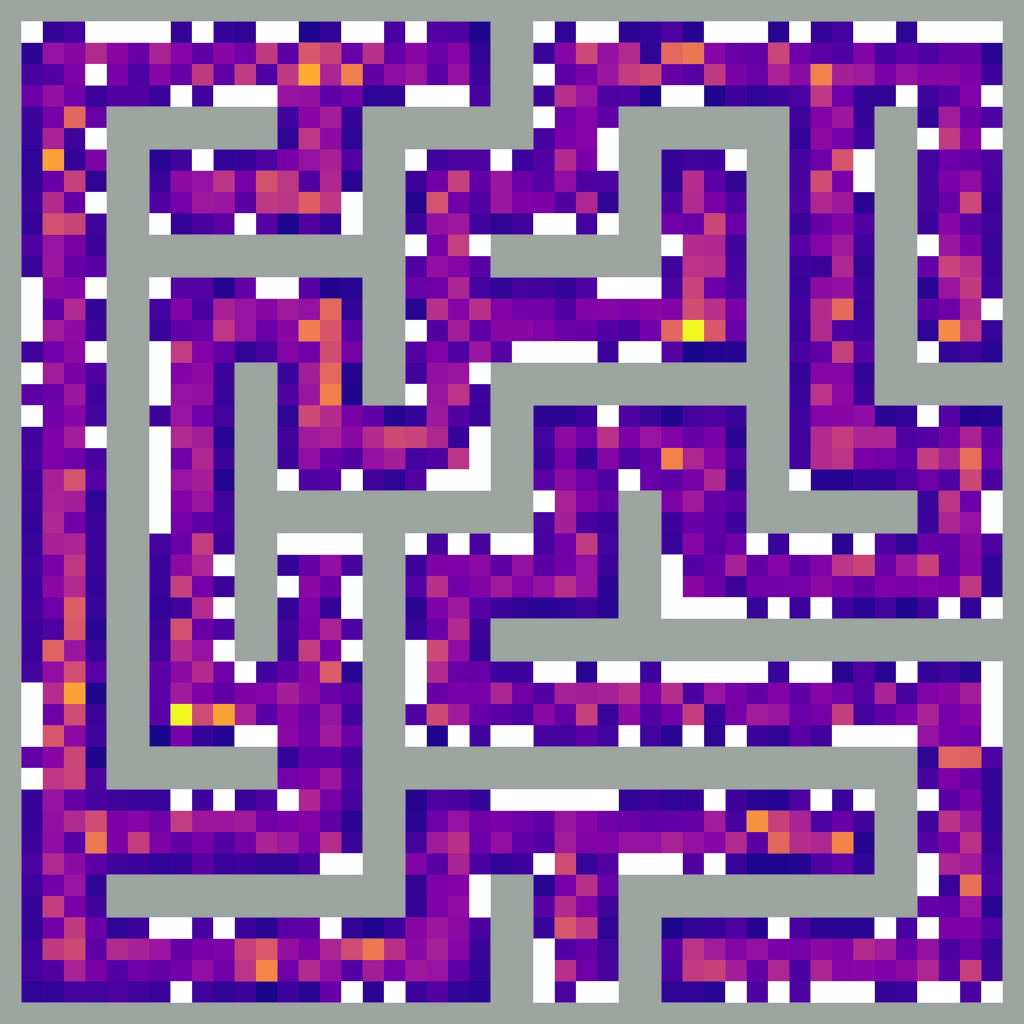}}%
    \hfill
    \subfigure[RRT]{\label{fig:results_comparison_progression_curved_rrt}\includegraphics[width=.12\textwidth]{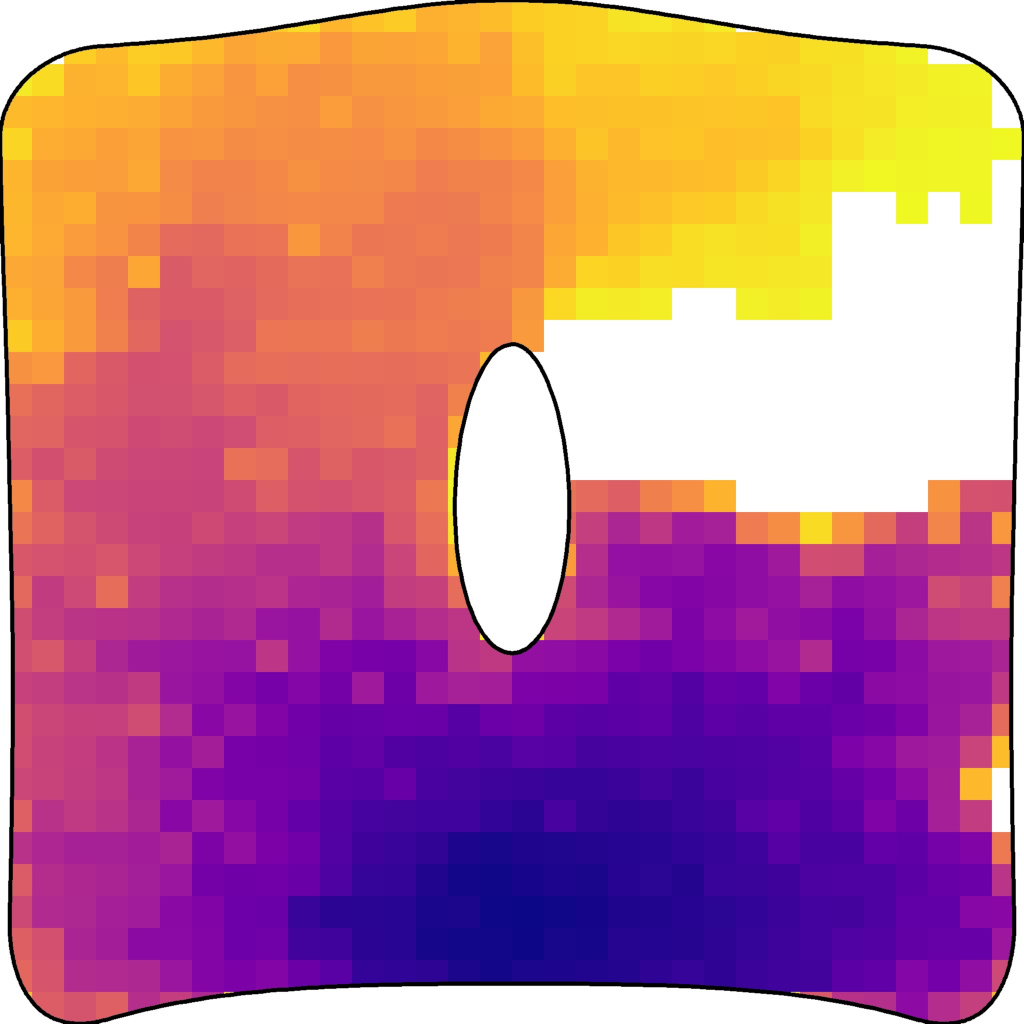}}%
    \hfill
    \subfigure[Biased]{\label{fig:results_comparison_progression_curved_biased}\includegraphics[width=.12\textwidth]{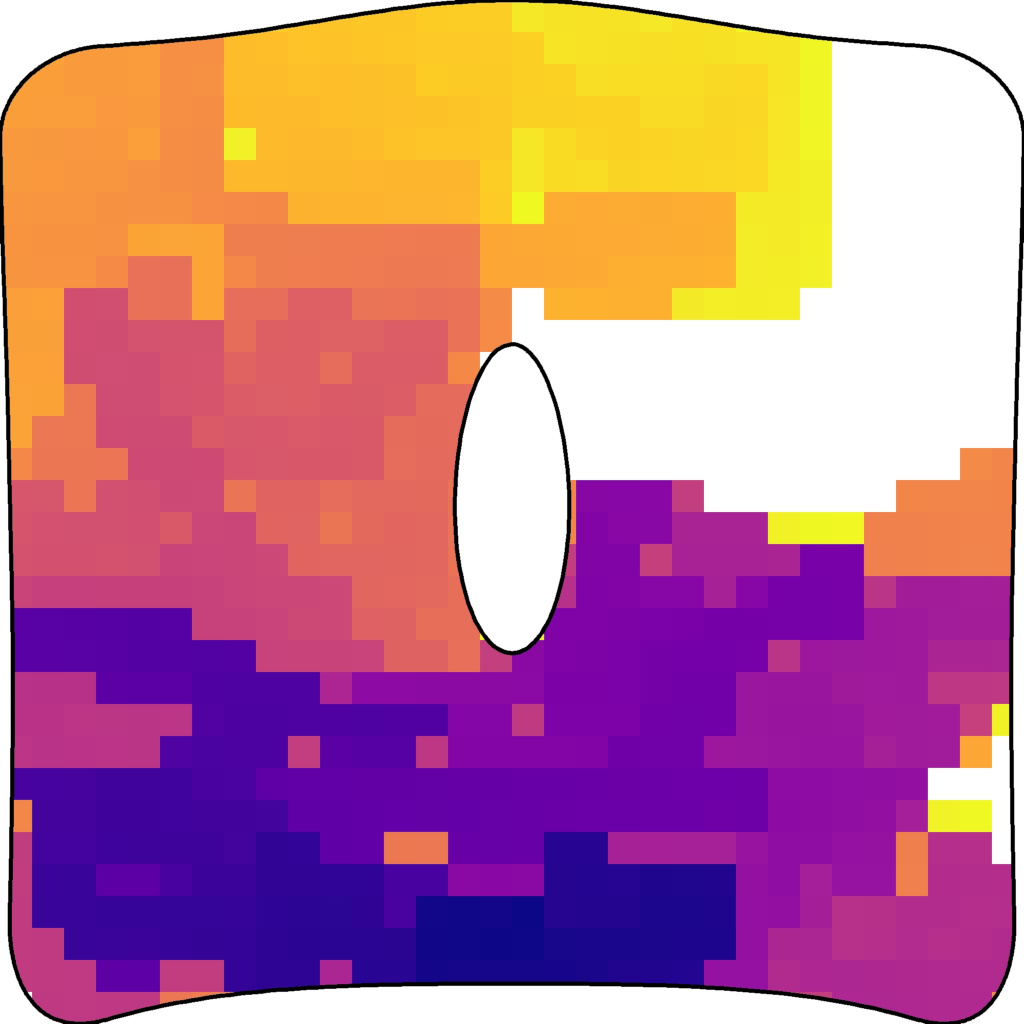}}%
    \hfill
    \subfigure[RRT]{\label{fig:results_comparison_progression_maze_rrt}\includegraphics[width=.12\textwidth]{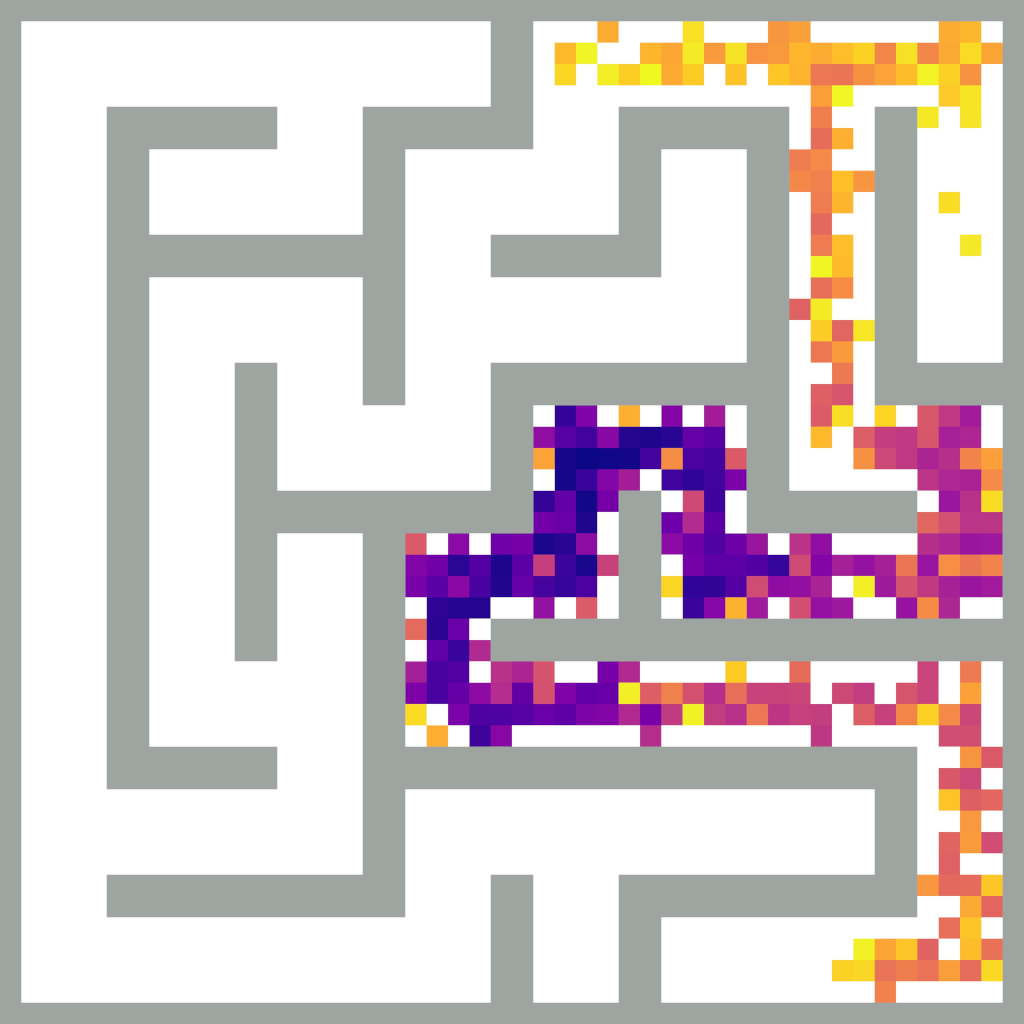}}%
    \hfill
    \subfigure[Biased]{\label{fig:results_comparison_progression_maze_biased}\includegraphics[width=.12\textwidth]{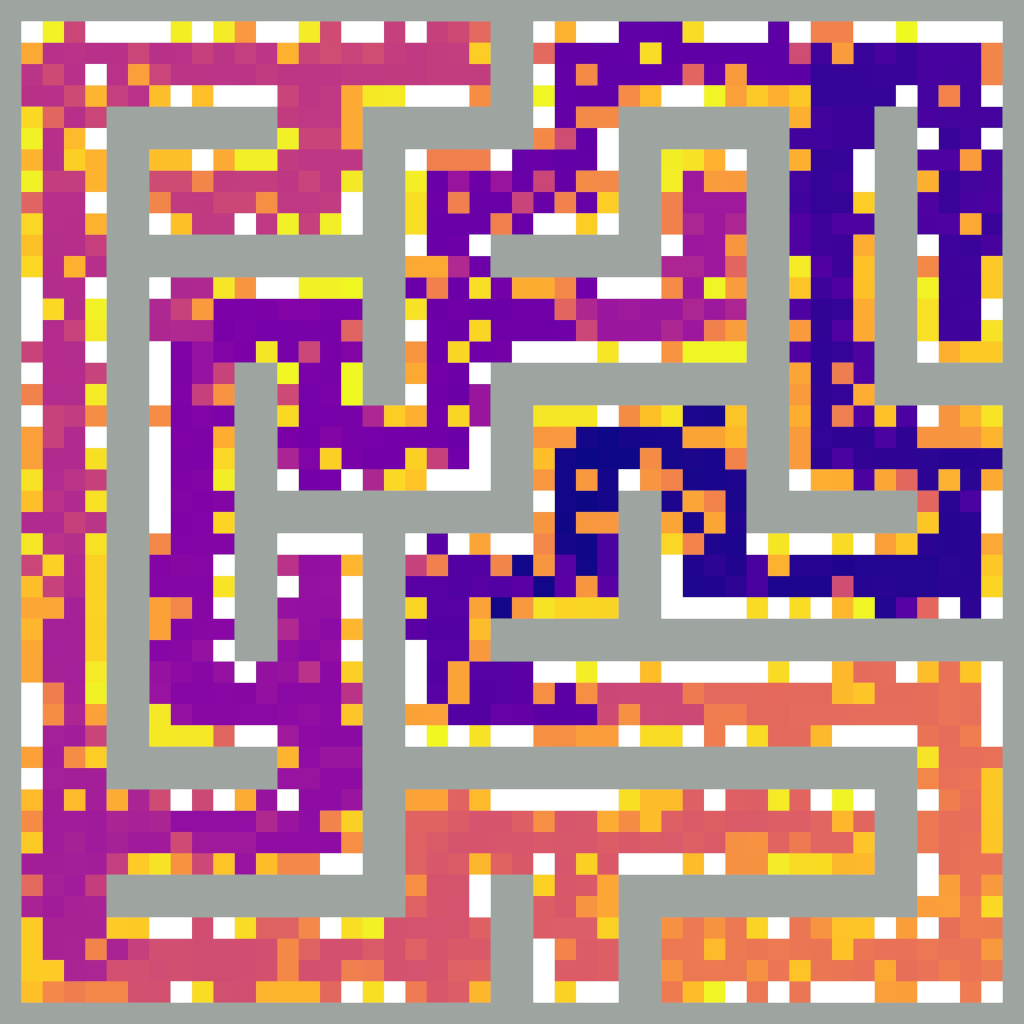}}\
    \caption{Sampling behavior of the Kuka robot in the curved and maze environments after \num{100000} samples. The four images on the left visualize the number of samples per cell, while the right images show the exploration order. Spots of high sampling density were manually annotated.}%
    \label{fig:results_comparison_qualitative}
\end{figure*}

The proposed approach was implemented using the constraint planning capabilities~\cite{kingston_exploring_2019} of the \emph{Open Motion Planning Library}~\cite{sucan_open_2012}.
We represent complex surfaces using the \emph{OpenCASCADE} CAD geometry kernel and utilize the \emph{MoveIt} framework~\cite{coleman_reducing_2014} for collision detection.
All benchmarks are performed on a desktop PC with an AMD Ryzen 9 9950X 16-core CPU running at a base frequency of \SI{4.3}{\giga\hertz} with 64 GB of DDR5 memory. %
As the evaluation primarily stresses a single CPU Core, we run up to five instances in parallel.

We start the evaluation by looking at the impact of the parameters \(d_{\mathrm{max}}\) and \(\sigma_{\mathrm{sample}}\) on runtime and coverage. %
Based on this, we select specific parameter values and analyze the sampling behavior and its impact on sample efficiency.
Finally, we analyze the required runtime to achieve accurate estimates.

\subsection{Sampling parameters}%
\label{subsec:results_parameters}

To investigate the effect of exploration parameters on the runtime and accuracy of the proposed approach, we use a test matrix composed of different robots and environments shown in Figure~\ref{fig:results_scenarios}.
For practical relevance, we use two robots with six degrees of freedom (Universal Robots UR5e, Kuka KR10 R900 sixx) and one with seven degrees of freedom (Franka Research 3).
For each of them, we evaluate three environments of different complexity.
In the simplest case, we have a symmetric surface with low curvature and without obstacles, focusing on the pure kinematic reachability.
In an advanced scenario, we have much higher surface curvatures and an additional obstacle that blocks transitions along one side of the surface.
Finally, an \(8\times8\) maze is used to study the effect of narrow passages.

All experiments in this chapter are run with \(\delta_{\mathrm{check}}=0.01\).
Unless otherwise noted, we use \(n_{\mathrm{grid}}=32\) for scenarios 1 and 2, and \(n_{\mathrm{grid}}=48\) for scenario 3.
As a baseline for the experiments, we performed exhaustive sampling of 10 million configurations to obtain a baseline of reachable cells for each robot and scenario.

We perform experiments with both proposed exploration approaches on each combination of robot and scenario.
For the RRT-based exploration approach, we use different values of \(d_{\mathrm{max}}\) within \([0.01, 0.19]\) in 10 steps of \(0.02\). %
For the biased exploration approach, we test \(\sigma_{\mathrm{sample}}\) values in \([0.01, 0.20]\) in steps of \(0.01\).
Each experiment was run 25 times with a timeout of \SI{60}{\second} per run.
The results of these experiments can be seen in Figure~\ref{fig:results_parameters}.

While the RRT-based and biased exploration perform comparably in the simple scenario with six degrees of freedom, biased exploration is significantly more performant and sample efficient in any more complex scenario.
In particular, the maze scenario is barely solvable using RRT exploration and will not be considered in further analysis.
As expected, a larger step or sample size generally leads to faster coverage.
With the increase, the probability of being unable to project a sample from a chart to \(\mathcal{M}\) also increases.
Together with our approach only registering the samples themselves and not the geodesics towards them, this increases the required sample time.
Across different robots, the behavior seems pretty similar, with the Franka robot achieving less coverage due to its higher-dimensional search space.
Both runtime and coverage across different scenarios differ quite a bit, which is to be expected due to the larger required joint motion in the curved scenario and the highly restricted environment in the maze scenario.
For this approach to be practical for real-world analysis of robot continuous reachability, we do, however, need to be able to use one set of parameters for any given surface and robot.
Based on the tradeoff between coverage efficiency and required runtime, we manually choose \(d_{\mathrm{max}} = 0.07\) for RRT-based exploration and \(\sigma_{\mathrm{sample}} = 0.04\) for biased exploration and will use these values for the remainder of the evaluation. %

\begin{figure*}[t]
    \includegraphics[width=\textwidth]{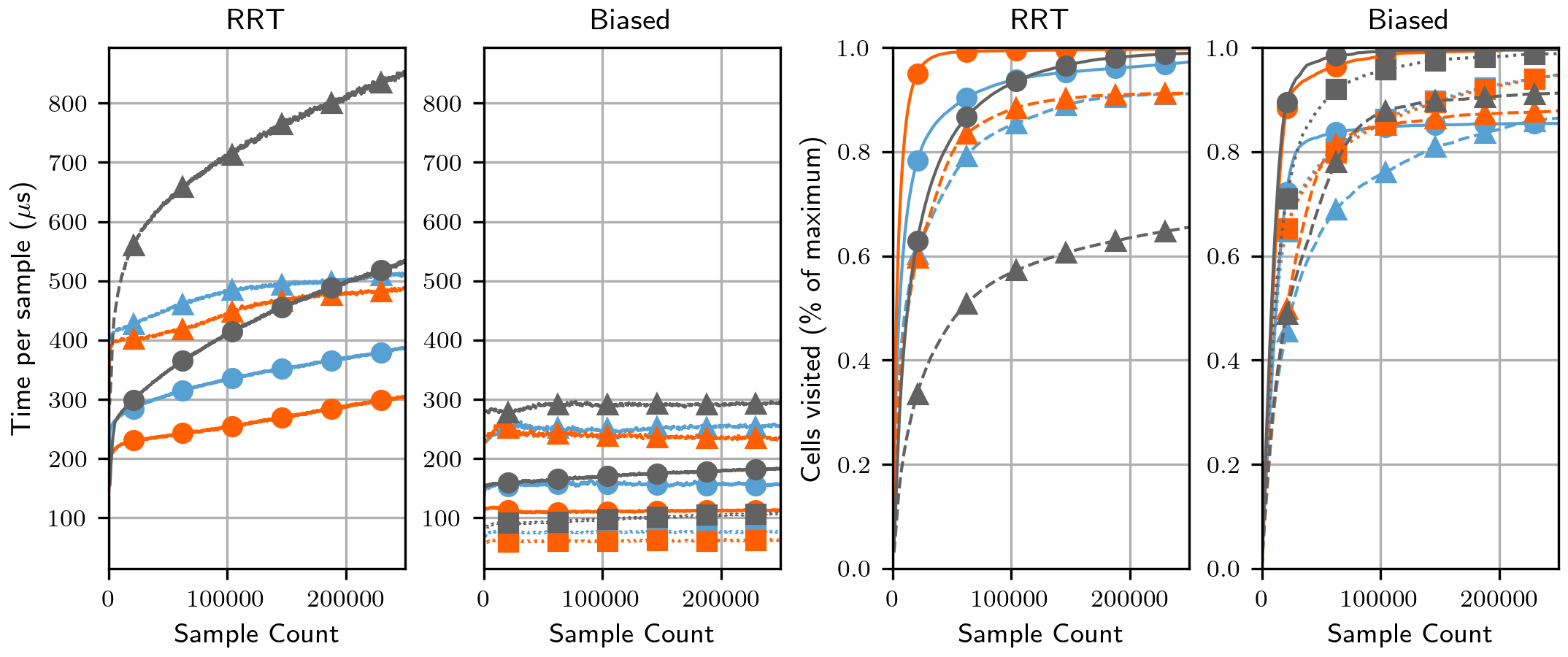}%
    \caption{Evaluation of mean sample time (left) and mean coverage (right) during exploration with \num{250000} samples. The sample times were smoothed using a bidirectional exponential moving average (EMA, \(\alpha=0.01\)).}%
    \label{fig:results_comparison}
\end{figure*}

\subsection{Sampling Behavior}%
\label{subsec:results_sampling}

In order to analyse the discrepancy in sampling performance between the RRT-based exploration and biased exploration, we ran exploration for \num{100000} samples for the Kuka robot in the curved and maze environments and observed the number of samples per cell on the surface and the order in which exploration occurred.
The results of this can be seen in Figure~\ref{fig:results_comparison_qualitative}.

First of all, we observe the density of samples per cell.
The biased approach leads to a generally uniform distribution without particular high spots.
Only the borders have a higher density, where the biased sampling tried to extend further and could not find valid new neighbor cells.
In contrast, one can clearly see spots of very high density for the RRT-based exploration.
Expansion of the RRT happens mostly uniformly in \(\mathcal{M}\), which does not translate to a uniform distribution on \(S\).
In the curved scenario that is especially visible, as the high-density region contains configurations close to a robot singularity, where steps in axes 4 and 6 will lead to minimal motion on the surface.

The expansion behavior is also visible in the expansion progression.
The RRT-based exploration gradually expands in all directions, without any discernible jumps in the resulting figure.
The biased sampling instead shows regions that were expanded rapidly, showing large jumps to adjacent regions that were only discovered later.
This is a clear result of the \(\frac{\log(\mathcal{I})}{S}\) scoring in~(\ref{eqn:approach_ranking}), which leads to newly covered cells being preferred for the next sampling steps.
In the maze environment in particular, one can see sections being explored at once before switching to a completely different part of the surface.

It is important to note that complete coverage of \(\mathcal{M}\) is computationally expensive and is often not required for an accurate estimate.
In the presence of obstacles and complex surfaces, it is, however, not generally possible to get accurate estimates without any exploration of the inherent kinematic redundancy of the problem.
We see this as the major contribution of the proposed concept and the main contribution compared to greedy search approaches in prior work~\cite{mcgovern_efficient_2021,mcgovern_general_2024}, which only perform a 2D grid search.

\begin{figure*}[t]
    \includegraphics[width=\textwidth]{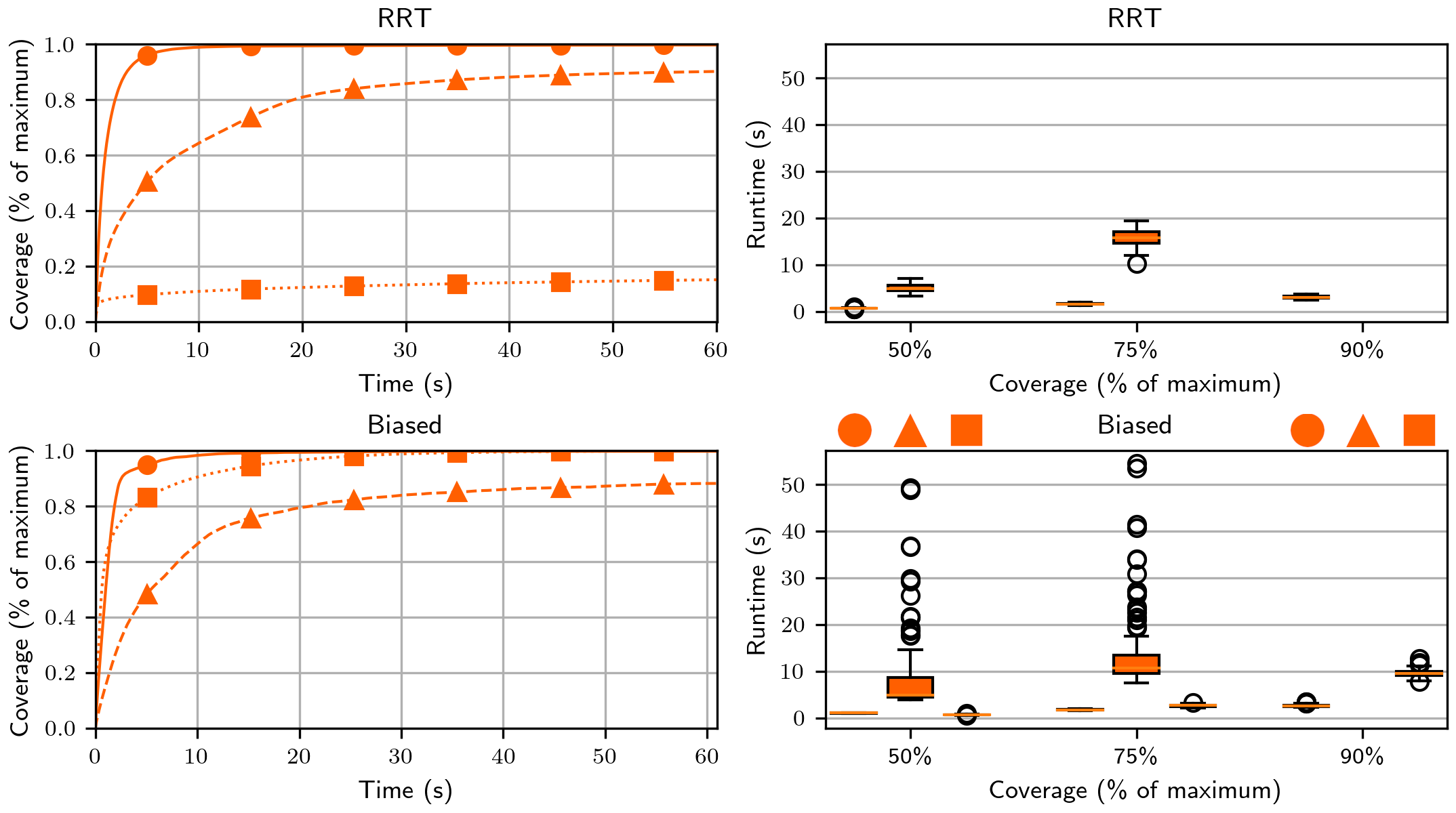}\
    \caption{Coverage over time and time to achieve a specific degree of coverage. All experiments were done \num{100} times, with the left plots showing the mean coverage at every point in time. Boxes for scenarios in which coverage to a certain degree was not consistently possible were omitted in the right plot.}%
    \label{fig:results_time}
\end{figure*}

\subsection{Performance}%
\label{subsec:results_performance}

As a final evaluation step, we analyse the performance of the proposed approach.
For this, we first look at sample time and coverage over the number of samples, before evaluating the average times to achieve coverage percentiles.

We run the exploration on all robots and scenarios for \num{100} iterations at \num{250000} samples each.
The resulting mean time per sample and mean coverage at each step can be seen in Figure~\ref{fig:results_comparison}.

The time taken to generate a sample for RRT-based exploration increases with the number of total samples, while the time taken for biased exploration seems mostly constant.
We attribute this mainly to the Nearest-Neighbor structure needed for RRT expansion, which is not required in the KPIECE weighting scheme.
This matches the more substantial time increase for the 7-DOF Franka arm in the RRT case, where an additional dimension needs to be tracked.
Comparing scenarios, the curved surface takes significantly more effort than the slightly rounded surface.
Perhaps counter-intuitively, the maze scenario (only considered for the biased case as discussed in Section~\ref{subsec:results_parameters}) takes an even smaller time per sample than the simple curved scenario.
The simple structure of the surface allows for very fast sampling, and due to the obstacle structure, the probability of discarding samples due to collisions stays approximately constant throughout the experiment.

Regarding coverage per sample, the RRT-based exploration seems to arrive at a high level of coverage with fewer samples than the biased approach.
This is, however, notably different for the 6-DOF Franka curved scenario, where a lot of nullspace motion is required to reach further parts of the surface.
Perhaps surprisingly, the simple scenario (especially for the UR robot) performs worse for biased sampling than RRT-based exploration.
We suspect this to be caused by the border of coverage on the surface, where the inherent IMACS biasing towards open charts leads to exhaustive coverage of all boundary cells.
At the same time, the biased exploration at some point no longer finds new cells near the border and thus searches in different parts.

While the prior steps have looked at estimation speed, they focused on per-sample performance and efficiency.
As a final step, we analyze the coverage estimate over the runtime of the approach.
For this, we run \num{100} iterations using the Kuka robot across all scenarios, the results of which can be seen in Figure~\ref{fig:results_time}.

Both approaches perform significantly better on the simple scenario than the curved one, reaching \SI{90}{\percent} coverage on average within the first three seconds.
The curved scenario requires more time, achieving \SI{80}{\percent} coverage only within 19 and 21 seconds.
Interestingly, the RRT-based exploration performs better in this case, but neither approach can consistently surpass \SI{90}{\percent} coverage in the allotted time.
For the lower curvature scenarios, biased exploration is faster, but has a much larger spread with significant execution time outliers, while RRT-based exploration performs more predictably.

\section{CONCLUSIONS AND FUTURE WORK}%
\label{sec:conclusion}
We have presented a novel approach for continuous coverage estimation using an implicit manifold constraint.
By transferring robot motions to an extended configuration space, we can capture the structure of valid motions and iteratively explore the complete robot action space.
Two sampling approaches are proposed and compared for its exploration and evaluated across a range of robots and scenarios.

We show that this general scheme can provide accurate estimates of the continuously reachable surface regions, which can be used as a building block for higher-level cell optimization or as a starting point for region-based tool path generation.
Its ability to work directly on complex CAD surfaces makes it applicable in real-world industrial contexts.
While its functionality is currently limited by its requirements on surface smoothness, straightforward extensions could project samples across discontinuities and expand the possible scope of application.
Furthermore, we plan to investigate further uses of manifold exploration for direct coverage path planning and enhanced robot performance metrics.

\balance{}
\bibliographystyle{IEEEtran}
\bibliography{Bibliography/IEEEabrv,Bibliography/bibliography}

\begin{thebibliography}{10}
\providecommand{\url}[1]{#1}
\csname url@rmstyle\endcsname
\providecommand{\newblock}{\relax}
\providecommand{\bibinfo}[2]{#2}
\providecommand\BIBentrySTDinterwordspacing{\spaceskip=0pt\relax}
\providecommand\BIBentryALTinterwordstretchfactor{4}
\providecommand\BIBentryALTinterwordspacing{\spaceskip=\fontdimen2\font plus
\BIBentryALTinterwordstretchfactor\fontdimen3\font minus \fontdimen4\font\relax}
\providecommand\BIBforeignlanguage[2]{{%
\expandafter\ifx\csname l@#1\endcsname\relax
\typeout{** WARNING: IEEEtran.bst: No hyphenation pattern has been}%
\typeout{** loaded for the language `#1'. Using the pattern for}%
\typeout{** the default language instead.}%
\else
\language=\csname l@#1\endcsname
\fi
#2}}

\bibitem{glorieux_coverage_2020}
E.~Glorieux, P.~Franciosa, and D.~Ceglarek, ``Coverage path planning with targetted viewpoint sampling for robotic free-form surface inspection,'' \emph{Robotics and Computer-Integrated Manufacturing}, vol.~61, p. 101843, 2020.

\bibitem{wen_uniform_2022}
Y.~Wen, D.~J. Jaeger, and P.~R. Pagilla, ``Uniform {Coverage} {Tool} {Path} {Generation} for {Robotic} {Surface} {Finishing} of {Curved} {Surfaces},'' \emph{IEEE Robotics and Automation Letters}, vol.~7, no.~2, pp. 4931--4938, 2022.

\bibitem{schneyer_segmentation_2023}
S.~Schneyer, A.~Sachtler, T.~Eiband, and K.~Nottensteiner, ``Segmentation and {Coverage} {Planning} of {Freeform} {Geometries} for {Robotic} {Surface} {Finishing},'' \emph{IEEE Robotics and Automation Letters}, vol.~8, no.~8, pp. 5267--5274, 2023.

\bibitem{li_five-axis_2023}
Z.~Li, K.~Tang, P.~Hu, and L.~Huang, ``Five-{Axis} {Trochoidal} {Sweep} {Scanning} {Path} {Planning} for {Free}-{Form} {Surface} {Inspection},'' \emph{IEEE Transactions on Automation Science and Engineering}, vol.~20, no.~2, pp. 1139--1155, 2023.

\bibitem{do_geometry-aware_2023}
V.-T. Do and Q.-C. Pham, ``Geometry-{Aware} {Coverage} {Path} {Planning} for {Depowdering} on {Complex} {3D} {Surfaces},'' \emph{IEEE Robotics and Automation Letters}, vol.~8, no.~9, pp. 5552--5559, 2023.

\bibitem{lin_robot_2017}
Y.-Y. Lin, C.-C. Ni, N.~Lei, X.~David~Gu, and J.~Gao, ``Robot {Coverage} {Path} planning for general surfaces using quadratic differentials,'' in \emph{2017 {IEEE} {International} {Conference} on {Robotics} and {Automation} ({ICRA})}, 2017, pp. 5005--5011.

\bibitem{mcgovern_uv_2022}
S.~McGovern and J.~Xiao, ``{UV} {Grid} {Generation} on {3D} {Freeform} {Surfaces} for {Constrained} {Robotic} {Coverage} {Path} {Planning},'' in \emph{2022 {IEEE} 18th {International} {Conference} on {Automation} {Science} and {Engineering} ({CASE})}, 2022, pp. 1503--1509.

\bibitem{jafari_surface_2020}
B.~H. Jafari and N.~Gans, ``Surface {Parameterization} and {Trajectory} {Generation} on {Regular} {Surfaces} {With} {Application} in {Robot}-{Guided} {Deposition} {Printing},'' \emph{IEEE Robotics and Automation Letters}, vol.~5, no.~4, pp. 6113--6120, 2020.

\bibitem{yang_improved_2025}
T.~Yang, J.~Valls~Miro, Y.~Wang, and R.~Xiong, ``An {Improved} {Maximal} {Continuity} {Graph} {Solver} for {Non}-{Redundant} {Manipulator} {Non}-{Revisiting} {Coverage},'' \emph{IEEE Transactions on Automation Science and Engineering}, vol.~22, pp. 3822--3834, 2025.

\bibitem{kingston_sampling-based_2018}
Z.~Kingston, M.~Moll, and L.~E. Kavraki, ``Sampling-{Based} {Methods} for {Motion} {Planning} with {Constraints},'' \emph{Annual Review of Control, Robotics, and Autonomous Systems}, vol.~1, no. Volume 1, 2018, pp. 159--185, 2018.

\bibitem{bordalba_randomized_2021}
R.~Bordalba, L.~Ros, and J.~M. Porta, ``\BIBforeignlanguage{English}{A {Randomized} {Kinodynamic} {Planner} for {Closed}-{Chain} {Robotic} {Systems}},'' \emph{\BIBforeignlanguage{English}{IEEE Transactions on Robotics}}, vol.~37, no.~1, pp. 99--115, Feb. 2021.

\bibitem{kingston_informing_2020}
Z.~Kingston, A.~M. Wells, M.~Moll, and L.~E. Kavraki, ``\BIBforeignlanguage{English}{Informing {Multi}-{Modal} {Planning} with {Synergistic} {Discrete} {Leads}},'' in \emph{\BIBforeignlanguage{English}{2020 {IEEE} {International} {Conference} on {Robotics} and {Automation} ({ICRA})}}, May 2020, pp. 3199--3205.

\bibitem{makhal_reuleaux_2018}
A.~Makhal and A.~K. Goins, ``Reuleaux: {Robot} {Base} {Placement} by {Reachability} {Analysis},'' in \emph{2018 {Second} {IEEE} {International} {Conference} on {Robotic} {Computing} ({IRC})}, 2018, pp. 137--142.

\bibitem{yao_enhanced_2024}
H.~Yao, R.~Laha, L.~F.~C. Figueredo, and S.~Haddadin, ``Enhanced {Dexterity} {Maps} ({EDM}): {A} {New} {Map} for {Manipulator} {Capability} {Analysis},'' \emph{IEEE Robotics and Automation Letters}, vol.~9, no.~2, pp. 1628--1635, 2024.

\bibitem{han_efficient_2021}
Y.~Han, J.~Pan, M.~Xia, L.~Zeng, and Y.-J. Liu, ``Efficient {SE}(3) {Reachability} {Map} {Generation} via {Interplanar} {Integration} of {Intra}-planar {Convolutions},'' in \emph{2021 {IEEE} {International} {Conference} on {Robotics} and {Automation} ({ICRA})}, 2021, pp. 1854--1860.

\bibitem{mcgovern_efficient_2021}
S.~McGovern and J.~Xiao, ``\BIBforeignlanguage{English}{Efficient {Feasibility} {Checking} on {Continuous} {Coverage} {Motion} for {Constrained} {Manipulation}},'' in \emph{\BIBforeignlanguage{English}{2021 {IEEE} 17th {International} {Conference} on {Automation} {Science} and {Engineering} ({CASE})}}, Aug. 2021, pp. 189--195.

\bibitem{mcgovern_general_2024}
------, ``A {General} {Approach} for {Constrained} {Robotic} {Coverage} {Path} {Planning} on {3D} {Freeform} {Surfaces},'' \emph{IEEE Transactions on Automation Science and Engineering}, vol.~21, no.~4, pp. 5546--5557, 2024.

\bibitem{kingston_exploring_2019}
Z.~Kingston, M.~Moll, and L.~E. Kavraki, ``\BIBforeignlanguage{English}{Exploring implicit spaces for constrained sampling-based planning},'' \emph{\BIBforeignlanguage{English}{The International Journal of Robotics Research}}, vol.~38, no. 10-11, pp. 1151--1178, 2019.

\bibitem{sucan_sampling-based_2012}
I.~A. Sucan and L.~E. Kavraki, ``A {Sampling}-{Based} {Tree} {Planner} for {Systems} {With} {Complex} {Dynamics},'' \emph{IEEE Transactions on Robotics}, vol.~28, no.~1, pp. 116--131, 2012.

\bibitem{jaillet_path_2013}
L.~Jaillet and J.~M. Porta, ``\BIBforeignlanguage{English}{Path {Planning} {Under} {Kinematic} {Constraints} by {Rapidly} {Exploring} {Manifolds}},'' \emph{\BIBforeignlanguage{English}{IEEE Transactions on Robotics}}, vol.~29, no.~1, pp. 105--117, Jan. 2013.

\bibitem{orthey_sampling-based_2024}
A.~Orthey, C.~Chamzas, and L.~E. Kavraki, ``\BIBforeignlanguage{English}{Sampling-{Based} {Motion} {Planning}: {A} {Comparative} {Review}},'' \emph{\BIBforeignlanguage{English}{Annual Review of Control, Robotics, and Autonomous Systems}}, vol.~7, May 2024.

\bibitem{sucan_open_2012}
I.~A. Şucan, M.~Moll, and L.~E. Kavraki, ``The {Open} {Motion} {Planning} {Library},'' \emph{IEEE Robotics \& Automation Magazine}, vol.~19, no.~4, pp. 72--82, Dec. 2012.

\bibitem{coleman_reducing_2014}
D.~Coleman, I.~A. Șucan, S.~Chitta, and N.~Correll, ``Reducing the {Barrier} to {Entry} of {Complex} {Robotic} {Software}: a {MoveIt}! {Case} {Study},'' \emph{Journal of Software Engineering for Robotics}, vol.~5, no.~1, pp. 3--16, May 2014.

\end{thebibliography}

\end{document}